\relax
\documentclass[letterpaper]{article} 

\usepackage{aaai2027}  

\usepackage{times}        
\usepackage{helvet}       
\usepackage{courier}      
\usepackage[hyphens]{url} 
\usepackage{graphicx}     
\usepackage{natbib}       
\usepackage{caption}      
\usepackage{algorithm}
\usepackage{algorithmic}
\usepackage{amsmath}
\usepackage{amssymb}
\usepackage{booktabs}     
\usepackage{multirow}
\usepackage{xcolor}
\usepackage{tikz}
\usetikzlibrary{arrows.meta,positioning}
\usepackage{pifont}
\usepackage{newfloat}
\usepackage{listings}
\DeclareCaptionStyle{ruled}{labelfont=normalfont,labelsep=colon,strut=off}
\floatstyle{ruled}
\newfloat{listing}{tb}{lst}{}
\floatname{listing}{Listing}

\newcommand{\method}{\textsc{ESV}}   
\newcommand{\cost}{\scriptsize}

\definecolor{pblue}{RGB}{31,97,160}
\definecolor{pgreen}{RGB}{0,122,87}
\definecolor{pgrey}{RGB}{95,95,95}

\newcommand{\mkopenci}[2]{\node[draw=#2,fill=white,line width=.6pt,circle,inner sep=1.5pt] at #1 {};}
\newcommand{\mkfillci}[2]{\node[draw=#2,fill=#2,circle,inner sep=1.5pt] at #1 {};}

\definecolor{okgreen}{RGB}{0,122,87}
\definecolor{okred}{RGB}{176,42,32}

\definecolor{cshared}{RGB}{226,226,226}
\definecolor{ccode}{RGB}{205,228,243}
\definecolor{ccall}{RGB}{252,230,192}
\definecolor{ccallb}{RGB}{191,118,0}
\definecolor{cout}{RGB}{206,238,227}
\definecolor{ctrap}{RGB}{250,222,218}


\title{Verification Beats Voting, If the Verifier Is Stronger:\\ Gold-Free Answer Selection Without Re-Solving}

\nocopyright

\begin{document}
\author{
    Ekansh Agrawal$^{1}$, Ritvij Sharma$^{2}$,
    Anand Ramaswamy Jayshree$^{3}$, Shanduojiao Jiang$^{4}$\\[2pt]
    {\normalfont\normalsize
    $^{1}$Texas A\&M University\\
    $^{2}$Starspark.ai\\
    $^{3}$University of Arizona\\
    $^{4}$Stanford University\\
    ekansh2023@tamu.edu, i-ritvij@starspark.ai, anandr@arizona.edu, sj99@stanford.edu}
}
\affiliations{}

\maketitle

\begin{abstract}
Self-consistency returns the majority answer over several chains of thought,
selecting by frequency alone, so it cannot see whether the reasoning behind a
candidate is valid. For small open-weight models the best answer already present
among four samples beats their majority vote by 6 points on GSM8K, 12 on
MATH-500 and 18 on JEEBench. We introduce \method{} (Explicit Strategy
Verification), a gold-free procedure that closes part of that gap without
re-solving: a zero-cost disagreement gate, one contrastive verification call
binding each candidate's strategy to quoted givens and discharging its
preconditions, and a deterministic rule accepting only candidates with no
mismatched binding and no violated condition. Across 2,918 instances \method{}
improves on CoT maj@4 by 5.0, 5.9 and 3.5 points (148 repairs against 4
regressions, $p \le 10^{-7}$ each), and by 3.8, 6.7 and 2.8 points on the
untouched originals alone, which no checker helped select. Universal
Self-Consistency reaches significance on none. Two experiments then locate that gain, and the second is a
negative result we report in full. The checker's identity decides the sign of
the effect: under the identical prompt, candidates and rule, the solver checking
its own samples instead loses 3.3, 2.3 and 2.2 points, and a likelihood-ratio
criterion fixed on development data predicts both outcomes in advance. But the
structured form is not what pays. Free-form adjudication by the same checker
beats \method{} by 0.6, 3.7 and 9.2 points, a shortfall we trace to the
checker's verdict rather than to our rule; what the form adds is an
interpretable failure label, on GSM8K only. Every problem is paired with a
structural perturbation that breaks or preserves the strategy the original
invites, manufacturing selection errors of known cause.
\end{abstract}

\section{Introduction}

Large language models solve multi-step reasoning problems by generating an
explicit chain of thought before committing to an answer~\cite{wei2022chain}. Because any single chain can go wrong, the standard
way to spend additional inference compute is to sample several of them and return their majority answer, a procedure known as self-consistency~\cite{wang2022self}. Majority voting selects by answer frequency alone, so it cannot see whether the reasoning behind a candidate is valid, and what that forfeits is substantial: for small open-weight models the best answer already present among four sampled chains beats their majority vote by six points on GSM8K, twelve on MATH-500 and eighteen on JEEBench. Gold-free remedies take one of two routes. One re-solves the problem from a critique of the first attempt~\cite{madaan2023self,shinn2023reflexion}; the other asks the model to nominate the most consistent of its own samples~\cite{chen2023universal}. The first pays full inference cost twice, and the second inherits the biases that produced the error.

A gold-free selection method has to satisfy three requirements at once, and no
existing method satisfies more than two: (1) it must never consult a reference
answer, so that it is deployable at inference time; (2) it must not re-solve the
problem, so that its cost is a fraction of the sampling budget it augments
rather than a multiple of it; and (3) it must not damage answers that were
already correct, since a method that fires on uncertainty can break a correct
majority as easily as repair a wrong one (Section~\ref{sec:requirements}).
Re-solving violates (2) by construction, and self-critique violates (3) as well,
because the blind spot that produced the error also confirms
it~\cite{huang2024large}. Universal Self-Consistency respects (1) and (2) but
asks the solver to adjudicate its own samples, and so gains little over the
majority vote it replaces.

We propose \textbf{\method{}} (\textbf{Explicit Strategy
Verification}), a multi-stage gold-free pipeline that selects among answers the
solver has already produced and never re-solves the problem
(Section~\ref{sec:method}): a disagreement gate that halts for free as soon as
the samples agree, so that verification is paid for only where the answer can
change; one call handing the distinct candidates, deduplicated and with their
vote counts hidden, to a checker stronger than the solver; and a rule that
either commits to a candidate or keeps the majority answer. Across GSM8K,
MATH-500 and JEEBench, with three open-weight solvers on the first and two on
the others, \method{} improves on majority voting by 5.0, 5.9 and 3.5 points,
against 0.9, 0.4 and $-1.9$ for Universal Self-Consistency, for 44\%, 83\% and
100\% of the sampling budget in extra compute. To attribute that gain we
evaluate each component on a benchmark of structural perturbations built for the
purpose, and the attribution is where the result turns: a free-form judge given
the same checker and the same candidates beats \method{} on all three. The gain
belongs to the checker, not to the verification format, and we report that
negative result as a finding rather than a footnote to the positive one.

Our contributions are:
\begin{itemize}
  \item We give a likelihood-ratio criterion for when a gold-free verdict may
        overturn a vote, fix it on development data, and show it predicts both
        the failure of self-checking and the success of an independent checker
        before either is measured.
  \item We construct a perturbation benchmark of matched problem pairs in which
        one variant breaks the strategy the original invites and the other
        preserves it, with an independently verified trap answer that makes the
        cause of each failure known.
  \item We show that the identity of the checker decides the sign of the effect.
        Holding the candidates, prompt and rule fixed, a stronger checker gains
        on every benchmark and the solver checking itself loses on every
        benchmark, extending known self-correction failures to a setting where
        nothing is re-solved.
  \item We report a negative result on verification format: \method{}'s
        structured bindings and conditions are beaten by free-form adjudication
        with the same checker on all three benchmarks. We trace the shortfall to
        the checker's verdict rather than to our decision rule, which a
        strictness ablation shows is not the binding constraint, and we find
        that the structured form's remaining value is interpretable error
        labels on GSM8K.
\end{itemize}

\section{Related Works}
\paragraph{Sampling and answer aggregation.}
Self-consistency samples several chains of thought and returns the modal answer
\cite{wang2022self}, and later work generalises the vote to free-form
outputs by asking the model itself to select the most consistent response
\cite{chen2023universal}. Agreement governs such aggregation completely: the
answer a plurality reaches is returned whether or not its derivation is sound,
so errors shared across samples are unrecoverable and a correct answer appearing
once can never win. We leave the sample budget untouched and replace the
frequency statistic with an explicit soundness check over the distinct
candidates, which lets a minority answer win when the majority's reasoning does
not survive verification.

\paragraph{Self-correction and its limits.}
Prior work on iterative refinement has models critique and revise their own
outputs \cite{madaan2023self, shinn2023reflexion}, while negative results find
that, with no external signal, language models cannot reliably identify errors
in their own reasoning \cite{huang2024large}. These studies establish \emph{that}
self-correction fails; they do not isolate the quantity responsible. Holding the
candidate set, the prompt and the decision rule fixed and varying only the
identity of the checker, we find a solver judges its own derivations sound an
order of magnitude more often than an independent checker does, and that this
miscalibration, rather than the verification format, turns self-checking into a
net loss.

\paragraph{Learned verifiers and LLM judges.}
A complementary line trains or prompts a separate model to assess candidate
solutions, as an outcome verifier \cite{cobbe2021training}, a step-level process
reward model \cite{lightman2024let}, or a general-purpose judge
\cite{zheng2023judging}. This work establishes that external scoring helps, but
conflates two factors: how capable the checker is, and the form in which it is
asked to deliberate. We separate them with a control holding the checker, its
candidates and its budget fixed while replacing structured verification with
free-form adjudication, and report that comparison in both directions, including
the regimes where the structured form is the weaker.

\paragraph{Perturbation and the memorization question.}
Robustness studies perturb benchmark problems to test whether apparent reasoning
survives distribution shift, by resampling symbolic templates, altering numeric
values, or inserting insignificant clauses \cite{mirzadeh2025gsm}. Such
perturbations stay largely at the surface and carry no annotation with respect to
the solution procedure, so a drop in accuracy cannot be traced to a specific
failure of transfer. We define perturbations relative to an explicit solution
strategy extracted from the original problem, and label each pair
strategy-breaking or strategy-preserving only after two independent verifiers
agree on the perturbed answer and confirm what that strategy does to it. For
breaking pairs we also record the trap answer the original strategy yields,
giving per-instance ground truth for the error mode a selector must detect.

\paragraph{Allocating test-time compute.}
Recent work treats inference-time computation as a resource to be scheduled,
trading additional samples, search, or verifier calls against accuracy \cite{snell2024scaling}. Our method is deliberately frugal within this framing: verification is gated on disagreement among the existing samples, so no
call is issued where the solver is already unanimous, and the checker is invoked
at most once per instance. We report the resulting cost in full, giving billed
and critical-path latency, solver and verifier tokens separately, and tokens per
net correction, so the accuracy gained can be weighed against drawing more
samples.

\section{Preliminaries and Problem Formulation}
\label{sec:prelim}

\paragraph{Sampling and candidates.}
Given a problem $x$ with reference answer $y$, a solver produces $n = 4$ chains
of thought and the answer each one reaches. Grouping answers that agree under a
type-aware normalization yields $K \le n$ distinct \emph{candidates}
$c_1, \dots, c_K$ with vote counts $v_1, \dots, v_K$ summing to $n$, each
carrying one representative solution. Self-consistency \cite{wang2022self}
returns the candidate with the most votes, which we write $m(x)$ and refer to
throughout as CoT maj@4, our baseline. We write CoT pass@4 for the oracle that
is correct whenever $y$ appears among the candidates.

\paragraph{The selection task.}
We study \emph{answer selection}: given the problem and the candidates the
solver has already produced, commit to one of them. Every method we compare
consumes the same four samples, drawn once per instance and reused unchanged,
so the comparison is about selection evidence rather than sampling luck. They
differ only in what they read off those samples: CoT maj@4 uses the vote
counts; Universal Self-Consistency \cite{chen2023universal} sees the samples
with duplicates shown, so multiplicity remains visible; the verification
methods see the problem text and the representative solutions with vote counts
withheld.

\subsection{Requirements on a gold-free selector}
\label{sec:requirements}

Three constraints separate a selector deployable at inference time from one
that merely explains a benchmark after the fact.

\begin{description}
\item[\textnormal{(R1) \emph{Gold-free.}}] The selector reads only the problem
  and the samples. The reference answer enters at scoring and nowhere else.
\item[\textnormal{(R2) \emph{No re-solving.}}] The selector issues no further
  call to the solver, so its cost is a fraction of the sampling budget it
  augments rather than a multiple of it.
\item[\textnormal{(R3) \emph{Non-destructive.}}] Because the selector may
  overturn $m(x)$, it must be judged by what it repairs \emph{and} what it
  damages, not by accuracy alone.
\end{description}

R3 fixes our unit of measurement. Against the maj@4 baseline, a \emph{fix} is
an instance the majority got wrong and the selector gets right, a \emph{break}
the reverse, and we report the two together with an exact McNemar test rather
than an accuracy delta on its own.

\paragraph{Selection is a two-sided bet.}
A selector constrained to the candidates cannot invent an answer the solver
never produced, which bounds both of its tails. Let $\mathcal{F}$ be the
instances where the majority is wrong but some candidate is right, and
$\mathcal{B}$ those where the majority is already right and the candidates
disagree. Over $N$ instances,
\begin{equation}
  -\frac{|\mathcal{B}|}{N}
  \;\le\; \Delta\mathrm{acc}
  \;\le\; \frac{|\mathcal{F}|}{N}
  \;=\; \mathrm{pass@}4 - \mathrm{maj@}4 ,
  \label{eq:bounds}
\end{equation}
since a unanimous-but-wrong instance contains no correct candidate and so
raises neither bound. The upper bound is the \emph{headroom} a selector
competes for; we report both sides in Section~\ref{sec:results}. R3 is forced
by Eq.~\eqref{eq:bounds} rather than adopted out of caution.

\paragraph{When a verdict is worth acting on.}
The vote is itself an informative signal, so a selector that overturns it needs
evidence strong enough to overcome that prior. Let $p(v)$ be the probability
that a candidate with $v$ votes is correct and
$\rho(v) = p(v)/(1 - p(v))$ the corresponding odds. A verifier returning
evidence $E$ about a candidate carries the likelihood ratio
$\Lambda = \Pr(E \mid \text{correct}) / \Pr(E \mid \text{wrong})$, and
preferring a $v_c$-vote candidate over a $v_m$-vote majority requires
\begin{equation}
  \Lambda \;>\; \rho(v_m) \,/\, \rho(v_c) .
  \label{eq:threshold}
\end{equation}
We estimate $p(v)$ on a development split of one benchmark and hold the
resulting thresholds fixed thereafter, so Eq.~\eqref{eq:threshold} states how
strong a checker must be before a verification call can pay for itself, fixed
before any verifier is measured rather than fitted to one afterwards.
Section~\ref{sec:results} compares the measured likelihood ratios against it.

\paragraph{Perturbation pairs.}
Rather than evaluate on problems whose failure modes are unknown, we evaluate
on matched pairs. Each item is a frozen pair $(P, P')$ together with the
solution strategy $S$ that the original invites, where $P'$ alters one
structure, relationship or constraint of $P$ while remaining well posed and
keeping the original answer format. In a \emph{strategy-breaking} pair,
applying $S$ unchanged to $P'$ gives a wrong answer, which we record as the
\emph{trap}; in a \emph{strategy-preserving} pair, $S$ still works but the
answer changes and must be recomputed. The preserving mode controls for mere
novelty, separating a method that degrades on any altered problem from one that
detects strategy misfit. A candidate equal to the trap is wrong for a known
reason, which gives per-instance ground truth for the error class that
frequency-based selection cannot see. Section~\ref{sec:setup} gives the
generation protocol and the dataset statistics.

\section{Method}
\label{sec:method}

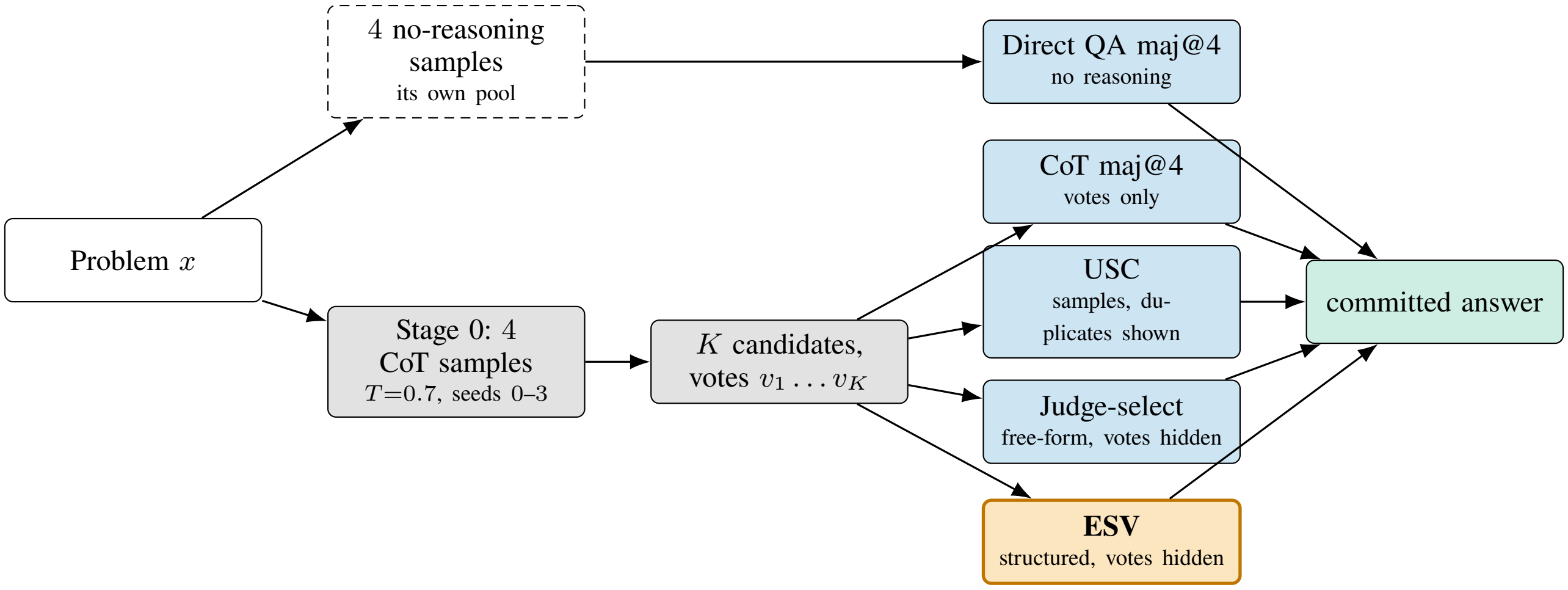
\begin{figure*}[t]
\centering
\begin{tikzpicture}[
  font=\small,
  base/.style={draw, rounded corners=2.5pt, align=center, inner sep=4pt,
               text width=25mm, minimum height=9mm},
  input/.style={base, fill=white},
  shared/.style={base, fill=cshared},
  own/.style={base, fill=white, densely dashed},
  sel/.style={base, fill=ccode},
  ours/.style={base, fill=ccall, draw=ccallb, line width=1pt},
  outbox/.style={base, fill=cout},
  ar/.style={-{Latex[length=2.2mm]}, semithick},
  cost/.style={font=\scriptsize},
]
\node[input]  (prob)  at (0,-0.2)     {Problem $x$};
\node[own]    (dpool) at (3.5,1.95)   {$4$ no-reasoning samples\\[-1pt]{\cost its own pool}};
\node[shared] (pool)  at (3.5,-1.3)   {Stage~0: $4$ CoT samples\\[-1pt]{\cost $T{=}0.7$, seeds $0$--$3$}};
\node[shared] (cands) at (7.0,-1.3)   {$K$ candidates, votes $v_1 \dots v_K$};

\node[sel]    (dqa)   at (10.6,1.95)  {Direct QA maj@4\\[-1pt]{\cost no reasoning}};
\node[sel]    (maj)   at (10.6,0.65)  {CoT maj@4\\[-1pt]{\cost votes only}};
\node[sel]    (usc)   at (10.6,-0.65) {USC\\[-1pt]{\cost samples, duplicates shown}};
\node[sel]    (jud)   at (10.6,-1.95) {Judge-select\\[-1pt]{\cost free-form, votes hidden}};
\node[ours]   (esv)   at (10.6,-3.25) {\textbf{\method{}}\\[-1pt]{\cost structured, votes hidden}};

\node[outbox] (ans)   at (14.1,-0.65) {committed answer};

\draw[ar] (prob)  -- (dpool);
\draw[ar] (prob)  -- (pool);
\draw[ar] (dpool) -- (dqa);
\draw[ar] (pool)  -- (cands);
\draw[ar] (cands) -- (maj);
\draw[ar] (cands) -- (usc);
\draw[ar] (cands) -- (jud);
\draw[ar] (cands) -- (esv);
\draw[ar] (dqa)   -- (ans);
\draw[ar] (maj)   -- (ans);
\draw[ar] (usc)   -- (ans);
\draw[ar] (jud)   -- (ans);
\draw[ar] (esv)   -- (ans);
\end{tikzpicture}
\caption{How the methods relate. One stage~0 pool (grey) is drawn per instance
and read unchanged by four of the five selectors, so the comparison is about
selection evidence rather than sampling luck. Direct QA is the exception: it
draws its own samples from a prompt that forbids reasoning (dashed), and is a
floor on recall rather than a selector. The four selectors differ only in what
they read off the pool. Every path commits to a candidate the solver already
produced, so CoT pass@4 is a hard ceiling throughout.}
\label{fig:overview}
\end{figure*}

We describe the method in the order it runs: how a perturbation pair is built
and frozen, the stages that turn four samples into one committed answer, then
the selectors we compare against. Figure~\ref{fig:overview} places all five
methods on the sampling step they share.

\subsection{Constructing the perturbation pairs}
\label{sec:construction}

One generator model produces each pair, and two verifier models have to agree
before it is accepted. The verifiers are independent of the generator and of
each other, and none of the three is a solver in any later experiment. One of
the two verifiers is, however, the same model we later use as the independent
checker, so every perturbed problem in the dataset is one that checker answered
correctly during construction. That is why we report the originals separately:
they carry no such acceptance filter, and the checker's gain over the
vote has to survive on them as well (Section~\ref{sec:results}).

\paragraph{Generation.}
Given an original problem $P$ and its reference answer, the generator names the
solution strategy $S$ that $P$ invites, alters exactly one condition, and
returns the perturbed problem $P'$ together with its new answer, the condition
it changed, the reason the change matters, an alternative strategy, the answer
$S$ would produce if applied unchanged, and a confidence level. A balancer
assigns each problem one mode, strategy-breaking or strategy-preserving, before
the first attempt and holds it across retries, which keeps the modes near parity
rather than letting the generator drift toward whichever is easier to produce.

\paragraph{Verification.}
Acceptance turns on two calls to each verifier, in an order chosen so that
anchoring is impossible.

\begin{description}
\item[\textnormal{\emph{Blind solve.}}] The verifier sees $P'$ alone, without
  $P$, without $S$ and without the generator's answer, and solves it. The pair is
  rejected if the verifier declares $P'$ ill-posed, if its answer disagrees
  with the generator's, or if its confidence falls below $0.6$. Withholding $P$
  is what makes the resulting answer an independent check rather than a
  ratification.
\item[\textnormal{\emph{Mechanical strategy application.}}] The verifier now
  sees $P$, $P'$, $S$ and the verified answer, and applies $S$ to $P'$ without
  adapting it. In breaking mode the pair is confirmed when the resulting answer
  differs from the verified answer; in preserving mode, when it agrees. A
  report that $S$ cannot be carried out at all counts as a soft confirmation of
  breaking.
\end{description}

A pair is frozen only if neither verifier objects and at least one returns a
concrete confirmation; a soft confirmation alone is never sufficient. Rejected
attempts are returned to the generator with the objection attached and retried
up to a cap, and every attempt is logged so that the frozen set can be rebuilt
from the log.

Consider one accepted breaking pair (GSM8K \#131). The perturbation moves
the comparison from Brian's seven-day total to his one-day distance, and the
original strategy still completes on $P'$ and still returns $35$, so $35$ is the
trap: any candidate equal to it is wrong for a reason recorded at construction
time.

\subsection{Explicit Strategy Verification}
\label{sec:esv}

\method{} adds three stages to a sampling step it shares with every other
method, and spends at most one model call on any instance.

\paragraph{Stage~0: shared sampling.}
Each solver answers each problem $n = 4$ times with chain-of-thought prompting
\cite{wei2022chain}, one call per sample at temperature $0.7$ under seeds $0$
to $3$, so that the four samples genuinely differ rather than repeating one
trajectory. Those four calls are the sampling budget against which every
later cost in this paper is measured. The samples are drawn once, written to
disk, and read unchanged by every selector, so no comparison below is affected
by sampling variance between methods.

\paragraph{Stage~1: the disagreement gate.}
The four answers are grouped into distinct candidates under a type-aware
normalization, in order of first appearance, with unreadable answers dropped.
If fewer than two candidates survive, the samples already agree, verification
has nothing to arbitrate, and the majority answer is committed with no call.
The gate is free and gold-free: it reads the samples only.

\paragraph{Stage~2: contrastive verification.}
Every surviving candidate is verified in a single call, each represented by one
of its solutions. The prompt states that the solutions reach different answers,
so at most one can be correct and all may be wrong; it presents them in
arbitrary order; and it withholds the vote counts while saying explicitly that
multiplicity is not evidence. For each candidate the checker returns the
strategy in one sentence, then one line per given the strategy depends on,
quoted word for word from the problem statement rather than from the solution's
restatement of it, marked \textsc{matches} or \textsc{mismatch}, then one
line per condition the strategy requires, marked \textsc{holds} or
\textsc{violated}. It then locates the first step at which the solutions
diverge, and finally returns \textsc{sound} or \textsc{flawed} per candidate
with a failure type, the first flawed step quoted verbatim, and the fact that
step should have asserted. Verification runs at temperature $0$. Asking for
quoted evidence before a verdict is the design choice under test. It is meant
to force the checker to commit to specific, checkable claims about the problem
text instead of judging fluency.

\paragraph{Stage~3: the decision rule.}
Selection is deterministic code with no further model call. Write $u_k$ for
candidate $k$'s verdict, $\mu_k$ for its count of \textsc{mismatch} bindings
and $\nu_k$ for its count of \textsc{violated} conditions. The candidate is
\emph{strictly sound} only if all three agree:
\begin{equation}
  \mathcal{S} = \{\, k : u_k = \textsc{sound},\; \mu_k = 0,\; \nu_k = 0 \,\}.
  \label{eq:strict}
\end{equation}
If $\mathcal{S}$ holds exactly one candidate, that candidate is committed. If it
holds several, the one with the most votes wins; ties go to the candidate
holding the majority answer $m(x)$, and then to the earliest in first-appearance
order. If it is empty, $m(x)$ is kept. The rule therefore never overturns the
baseline on evidence it does not have: where verification fails to discriminate,
\method{} returns what maj@4 would have returned, and the vote ranks only the
candidates verification has already cleared. Requiring the verdict to be
consistent with the checker's own line-by-line findings costs nothing and
discards verdicts the checker has already contradicted.

\paragraph{Two consequences of this shape.}
\method{} selects only among answers the solver already produced, so CoT pass@4
is a hard ceiling and no new answer can be introduced. And because an empty
$\mathcal{S}$ returns the majority answer, \method{} is a filter on
self-consistency rather than a replacement for it: it can only act where the
samples already disagree, and declines by default.

\paragraph{Checker conditions.}
We run stage~2 in two conditions that differ in nothing but which model is
called. Under \emph{self-check} the solver verifies its own samples; under
\emph{independent check} a single fixed checker verifies the samples of every
solver. Everything else (prompt, candidates, order, temperature, decision
rule) is held constant, so the contrast isolates the verifier.

\subsection{Methods compared}
\label{sec:comparisons}

Every selector consumes the stage~0 samples unchanged and differs only in the
evidence it reads off them.

\noindent\textbf{Direct QA} is the one arm that does not. It draws its own four
samples from a prompt that forbids reasoning and takes their majority, which
shows that the benchmarks are not solvable by recall.

\noindent\textbf{CoT pass@1} is the accuracy of a single sample, and
\textbf{CoT maj@4} \cite{wang2022self} the majority over four, which is our
baseline and the answer every selector must either confirm or overturn. \textbf{CoT
pass@4} marks an instance correct whenever any candidate is correct, giving the
oracle ceiling of Eq.~\eqref{eq:bounds}.

\noindent\textbf{Universal Self-Consistency} \cite{chen2023universal} is the
prior gold-free selector and our principal comparison. It receives all four
samples with duplicates shown, so consensus remains visible, and names the
sample it selects. Because an unresolved selection silently falls back to the
majority answer, we report the rate at which USC resolves a selection before
reporting its accuracy.

\noindent\textbf{Judge-select} is a form control, not a baseline. It is matched
to stage~2 on the checker, the candidates, their order, the withheld vote
counts, and the license to reject every candidate, and it reasons freely before
naming its pick. Two of those choices deliberately favor the control: free-form
reasoning makes the comparison structured-versus-unstructured rather than
reasoning-versus-none, and the license to decline denies \method{} a free win on
instances where no candidate survives. Judge-select therefore isolates what
\method{}'s structured form contributes once the checker's raw capability is
held fixed.

Reference answers enter at scoring and nowhere else, for every method.

\section{Experimental Setup}
\label{sec:setup}

\paragraph{Datasets.}
We build pairs from GSM8K \cite{cobbe2021training}, MATH-500 \cite{hendrycks2021math,lightman2024let} and the mathematics questions of JEEBench \cite{arora2023jeebench}, sampling 200, 250 and 200 source problems with a fixed seed and running the protocol of Section~\ref{sec:method}. This freezes 198, 244 and 193 pairs, split almost evenly between breaking and preserving modes, with acceptance rates of 95--100\% per mode. MATH-500 keeps every answer format, including 88 symbolic expressions, which one normalizer scores for every method. Each pair contributes two instances, $P$ and $P'$, per solver. On GSM8K we chose the decision rule and the checker on pairs 0--59 (357 instances) and evaluated them once on pairs 60--197 (813 held-out instances), against a go/no-go criterion fixed beforehand: ESV must beat CoT maj@4 by at least 3 points and must beat USC. Six GSM8K originals with wrong or underdetermined published answers are excluded from every score. JEEBench served as the development benchmark for an earlier version of the pipeline, so we treat it as a third data point rather than held-out evidence and mark it as such in every table. Judge-select was designed after the held-out run. The clean test is the held-out GSM8K column, with MATH-500 as replication and JEEBench as a stress case.

\paragraph{Models.}
The solvers are Qwen2.5-7B-Instruct \cite{qwen2025qwen25}, Mistral-NeMo-12B and Gemma-3-4B-it \cite{gemma2025gemma3}, small enough that their majority vote leaves 6.1, 12.4 and 17.9 points of headroom below pass@4. Larger solvers leave 2.1 points on GSM8K and almost nothing to select. MATH-500 and JEEBench were run on Mistral-NeMo and Gemma-3-4B only. The independent checker, shared by all solvers, is gpt-oss-120b \cite{openai2025gptoss}, a mixture-of-experts reasoning model with 116.8B total and 5.1B active parameters. Pairs were generated by \texttt{deepseek/deepseek-v4-flash-0731} and accepted by \texttt{qwen/qwen3.5-35b-a3b} and \texttt{openai/gpt-oss-120b}; we name every model by the identifier OpenRouter served, since that is what the stored responses record and what a reproduction must request. The first two have no technical report at the time of writing, so we cite the served identifier rather than a paper. The checker also accepted the pairs, so Section~\ref{sec:analysis} reports originals separately.

\paragraph{Baselines.}
Section~\ref{sec:method} defines the compared methods. Re-solving methods \cite{madaan2023self,shinn2023reflexion} violate R2 and are excluded.

\paragraph{Metrics.}
Against maj@4 on the same instances we count fixes $F$ and breaks $B$ (at most $|\mathcal{F}|$ and $|\mathcal{B}|$), report $\Delta = (F-B)/N$ with a 95\% paired interval, and test $F$ against $B$ with an exact two-sided McNemar test \cite{mcnemar1947}. Checker quality is measured per candidate: TPR is the share of correct candidates judged strictly \textsc{sound}, FPR the share of wrong ones, and $\Lambda = \mathrm{TPR}/\mathrm{FPR}$ is compared against Eq.~\eqref{eq:threshold}. Cost is prompt plus completion tokens and critical-path latency (the four samples run concurrently; the extra call waits for them). The tests treat instances as independent, although instances from one source problem are correlated across solvers, so on GSM8K we also report each solver separately.

\paragraph{Implementation details.}
Nothing is trained; all models are queried through the OpenRouter API. Solvers draw four samples at temperature 0.7 with seeds 0--3; the checker, judge-select and USC run at temperature 0, with gpt-oss-120b at reasoning effort medium. Numeric answers match within 0.01, and we draw one set of samples per instance, so seed variance is not measured. Frozen data, all raw responses and the analysis code are released at \url{https://anonymous.4open.science/r/Project_structural-perturbations-esv-09E7/}; every table regenerates offline from the stored responses.

\section{Results}
\label{sec:results}

\begin{table*}[t]
\centering
\small
\setlength{\tabcolsep}{4.2pt}
\begin{tabular}{lc rr rr rr}
\toprule
 & & \multicolumn{2}{c}{GSM8K (1{,}170; 3 solvers)} & \multicolumn{2}{c}{MATH-500 (976; 2 solvers)} & \multicolumn{2}{c}{JEEBench$^{\ddagger}$ (772; 2 solvers)} \\
\cmidrule(lr){3-4}\cmidrule(lr){5-6}\cmidrule(lr){7-8}
Method & Calls & Acc.\ (\%) & $F$ / $B$ & Acc.\ (\%) & $F$ / $B$ & Acc.\ (\%) & $F$ / $B$ \\
\midrule
Direct QA maj@4 & 4$^\dagger$ & 12.1 & 7 / 900$^{***}$ & 10.7 & 22 / 468$^{***}$ & 11.9 & 57 / 150$^{***}$ \\
CoT pass@1 & 1 & 85.1 & & 51.6 & & 20.5 & \\
CoT maj@4 (baseline) & 4 & 88.5 & & 56.4 & & 24.0 & \\
Universal Self-Consistency & 4 + 1 & 89.3 & 19 / 9 & 56.8 & 34 / 30 & 22.0 & 32 / 47 \\
\midrule
ESV, self-check & 4 + 1 & 85.1 & 18 / 57$^{***}$ & 54.1 & 28 / 50$^{*}$ & 21.8 & 33 / 50 \\
ESV, gpt-oss-120b & 4 + 1 & 93.5 & 61 / 2$^{***}$ & 62.3 & 59 / 1$^{***}$ & 27.5 & 28 / 1$^{***}$ \\
Judge-select, gpt-oss-120b & 4 + 1 & \textbf{94.1} & 67 / 1$^{***}$ & \textbf{66.0} & 99 / 5$^{***}$ & \textbf{36.7} & 98 / 0$^{***}$ \\
\midrule
CoT pass@4 (oracle) & 4 & 94.5 & 71 / 0 & 68.8 & 121 / 0 & 41.8 & 138 / 0 \\
\bottomrule
\end{tabular}
\caption{Accuracy on the perturbation sets. $F$/$B$: fixes and breaks against CoT maj@4 on the same instances; exact McNemar $^{*}p<0.05$, $^{***}p<0.001$. ``4 + 1'': one extra call, issued only when the samples disagree. Judge-select is the form control (Section~\ref{sec:method}). Bold: best non-oracle method. $^\dagger$Direct QA draws its own four samples without reasoning. $^{\ddagger}$JEEBench is a development benchmark rather than held-out evidence: it shaped an earlier version of this pipeline (Section~\ref{sec:setup}), so its column is a third data point and not confirmation. Half of each benchmark is perturbed problems that gpt-oss-120b helped accept, so those instances are filtered by the checker's own competence; Section~\ref{sec:analysis} reports the untouched originals, which passed through no such filter.}
\label{tab:main}
\end{table*}

\subsection{Main Results}
\label{sec:main}

With gpt-oss-120b as checker, ESV improves on CoT maj@4 by 5.0, 5.9 and 3.5 points (Table~\ref{tab:main}; 61/2, 59/1 and 28/1 fixes against breaks, $p \le 10^{-7}$ each) and beats USC head to head by 53/4, 71/17 and 69/27. It passed the criterion fixed before the held-out run, reaching 94.1\% on the 813 held-out GSM8K instances against 88.8\% for maj@4 and 89.7\% for USC, and every GSM8K solver gains on its own (Qwen 91.0 to 94.6\%, NeMo 84.4 to 92.8\%, Gemma 90.0 to 93.1\%). USC does not significantly beat the majority vote on any benchmark ($p = 0.087$, $0.71$, $0.11$), and its 9 unresolved selections out of 303 on GSM8K (0 of 601 on MATH-500, 20 of 694 on JEEBench) are too few to explain the null result. Self-check ESV lowers accuracy on all three benchmarks ($-3.3$, $-2.3$ and $-2.2$ points; $p = 7\times10^{-6}$, $0.017$, $0.078$). Since self-check and the gpt-oss-120b configuration share the prompt, the candidates and the rule, the checker alone turns a loss into a gain. This extends a known self-correction failure \cite{huang2024large} to a setting with no re-solving.

The form control shows where that gain comes from. Judge-select gives the same checker the same candidates without ESV's bindings, conditions or verdict form, and it is more accurate than ESV on every benchmark, by 0.6, 3.7 and 9.2 points (paired 7/0, 40/4 and 73/2; $p = 0.016$, $2\times10^{-8}$, $2\times10^{-19}$). The GSM8K gap turns on
7 paired disagreements out of 1{,}170 instances and comes from a single sample
pool, so we read its sign but not its size as settled; the MATH-500 and
JEEBench gaps rest on 44 and 75 paired disagreements. Net of breaks it recovers 93, 78 and 71\% of the oracle headroom where ESV recovers 83, 48 and 20\%. The accuracy comes from putting a stronger checker in front of the solver's own candidates; ESV's structured form contributes nothing to it and costs accuracy on harder problems.

\subsection{Ablation Study}
\label{sec:ablation}

Table~\ref{tab:ablation} isolates each component. A cascade that keeps ESV and defers to judge-select only when ESV abstains recovers almost all of the difference but never beats judge-select alone (paired 0/2, 2/3, 2/2). Where ESV's rule commits to one candidate the two methods agree: they differ on 2 of 231 such GSM8K instances and on none of 187 on MATH-500. The structured form therefore decides only where ESV abstains. Without the gate, ESV's GSM8K overhead would rise from 43.6\% to 168\% extra tokens.

The lower block tests whether our own rule holds \method{} back, since
Eq.~\eqref{eq:strict} demands a \textsc{sound} verdict with zero
\textsc{mismatch} and zero \textsc{violated} lines. It does not. Those extra
conditions almost never reject a candidate on their own: a \textsc{sound}
verdict arrives with no flagged line in 231 of 231 verified GSM8K candidates,
269 of 271 on MATH-500 and 79 of 79 on JEEBench, so accepting on the verdict
alone, or tolerating one flagged line, reproduces \method{} exactly. Only a rule
that can commit where nothing is \textsc{sound} changes anything, and ranking by
fewest flags is worse everywhere. The conservatism is in the checker's verdict,
not in the rule that reads it.

\begin{table}[t]
\centering
\small
\setlength{\tabcolsep}{3pt}
\begin{tabular}{lrrr}
\toprule
$\Delta$ vs.\ maj@4 (points) & GSM8K & MATH & JEE \\
\midrule
ESV, gpt-oss-120b checker & +5.0 & +5.9 & +3.5 \\
\quad checker $\to$ the solver itself & $-3.3$ & $-2.3$ & $-2.2$ \\
\quad structured form $\to$ free-form & \textbf{+5.6} & \textbf{+9.6} & \textbf{+12.7} \\
\quad + free-form fallback on abstain & +5.5 & +9.5 & \textbf{+12.7} \\
\midrule
\multicolumn{4}{@{}l}{\textit{decision rule, Eq.~\eqref{eq:strict}}} \\
\quad accept on verdict alone & +5.0 & +5.9 & +3.5 \\
\quad tolerate one flagged line & +5.0 & +5.9 & +3.5 \\
\quad rank by fewest flagged lines & +3.0 & +3.6 & +2.8 \\
\bottomrule
\end{tabular}
\caption{One change at a time from \method{} with the gpt-oss-120b checker;
free-form is judge-select. Each row changes exactly one component and leaves the
shared stage~0 samples, the candidate text and the scoring untouched. Removing
the gate leaves accuracy unchanged by construction, since a unanimous instance
offers nothing to select. The three rule variants relax
Eq.~\eqref{eq:strict}: the first two leave the result unchanged because a
\textsc{sound} verdict almost never carries a flagged line, and the third, which
is the only variant that can commit where no candidate is \textsc{sound}, is
worse everywhere.}
\label{tab:ablation}
\end{table}

\subsection{Analysis}
\label{sec:analysis}

\begin{figure}[t]
\centering
\begin{tikzpicture}[x=1.93cm, y=0.46cm, font=\footnotesize]
\fill[black!8] (-0.06,0.45) rectangle (0.987,6.60);
\draw[black!35] (0,0.45) -- (0,6.60)
  node[anchor=south, font=\scriptsize, black!55] {$1.0$};
\draw[black!60, densely dashed] (0.987,0.45) -- (0.987,6.60)
  node[anchor=south, font=\scriptsize] {$9.7$};
\draw[black!60, densely dashed] (1.511,0.45) -- (1.511,6.60)
  node[anchor=south, font=\scriptsize] {$32.4$};
\draw[black!55] (-0.06,0.45) -- (2.62,0.45);
\foreach \x/\lab in {0/1, 0.477/3, 1/10, 1.477/30, 2/100, 2.477/300}
  \draw[black!55] (\x,0.45) -- (\x,0.30) node[below=0pt, font=\scriptsize] {$\lab$};
\node[font=\scriptsize] at (1.28,-0.95) {likelihood ratio of \textsc{sound} (log scale)};
\newcommand{\lrrow}[4]{%
  \node[anchor=east, font=\scriptsize] at (-0.10,#1) {#2};
  \draw[black!45, line width=.5pt] (#3,#1) -- (#4,#1);}
\lrrow{6}{GSM8K, gpt-oss-120b}{1.677}{2.545}
  \mkfillci{(1.677,6)}{pblue}  \mkopenci{(2.545,6)}{pblue}
\lrrow{5}{GSM8K, self}{0.204}{0.362}
  \mkfillci{(0.204,5)}{ccallb} \mkopenci{(0.362,5)}{ccallb}
\lrrow{4}{MATH-500, gpt-oss-120b}{1.072}{1.455}
  \mkfillci{(1.072,4)}{pblue}  \mkopenci{(1.455,4)}{pblue}
\lrrow{3}{MATH-500, self}{0.114}{0.230}
  \mkfillci{(0.114,3)}{ccallb} \mkopenci{(0.230,3)}{ccallb}
\lrrow{2}{JEEBench, gpt-oss-120b}{1.689}{1.812}
  \mkfillci{(1.689,2)}{pblue}  \mkopenci{(1.812,2)}{pblue}
\lrrow{1}{JEEBench, self}{0.021}{0.037}
  \mkfillci{(0.021,1)}{ccallb} \mkopenci{(0.037,1)}{ccallb}
\end{tikzpicture}
\caption{Likelihood ratio of a \textsc{sound} verdict (filled) against the
Eq.~\eqref{eq:threshold} bars for the vote splits four samples allow (dashed): $32.4$ for 3--1,
$9.7$ for 2--1--1, and $1.0$ for a 2--2 tie. Five votes do not fit in four
samples, so no bar exists for a 2-vote challenger against a 3-vote majority.
Open circles add \textsc{flawed} on a competing candidate, assuming
independence, which makes them an upper bound rather than an estimate. The
self-checker (orange) never leaves the shaded region, where no
untied vote can be overturned, at either marker; the independent checker (blue)
clears the 2--1--1 bar everywhere on the filled marker alone, so that
conclusion does not rest on the independence assumption.}
\label{fig:lr}
\end{figure}
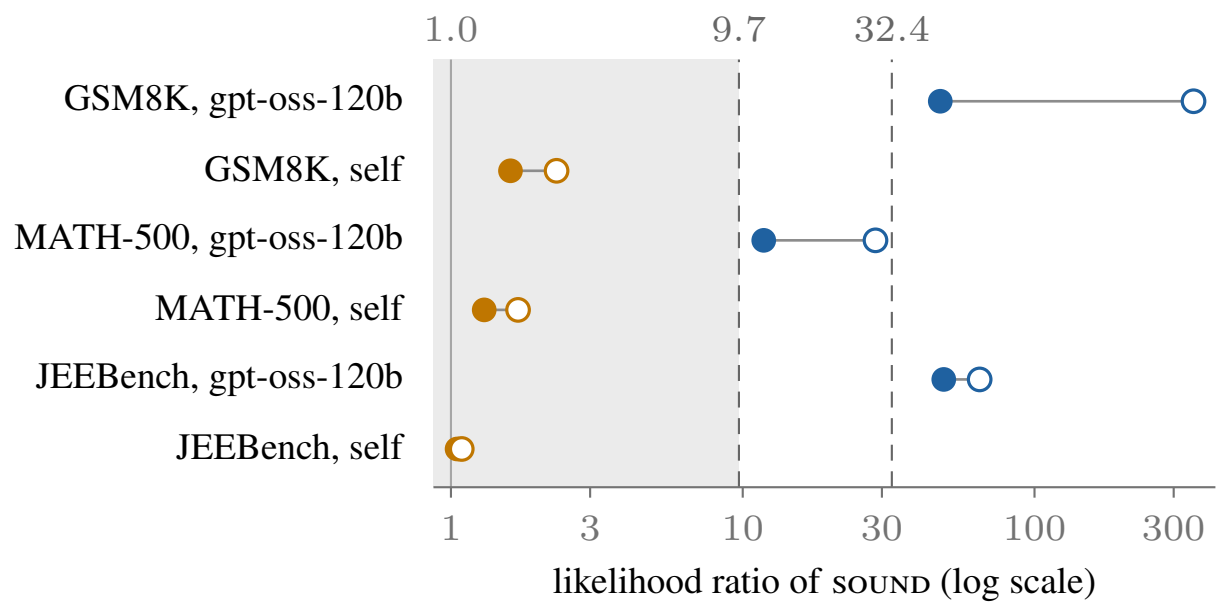

\paragraph{Why self-verification fails.}
On the GSM8K development split a candidate with one, two or three of four votes is correct 9.7, 51.1 and 77.8\% of the time. By Eq.~\eqref{eq:threshold}, overturning a 3--1 split then needs $\Lambda > 32.4$ and lifting a single-vote candidate in a 2--1--1 split needs $\Lambda > 9.7$; a tie needs only $\Lambda > 1$. The self-checker's \textsc{sound} verdict carries $\Lambda = 1.6$ on GSM8K and 1.3 on MATH-500, and even \textsc{sound} on the challenger with \textsc{flawed} on the majority reaches only 2.3 and 1.7 if the verdicts are independent (Figure~\ref{fig:lr}). That independence is an approximation, and it is the generous one: both verdicts come from a single call on a prompt that states at most one candidate can be correct, and both rest on one reading of the givens, so their errors are positively correlated. Were they perfectly correlated the second verdict would be implied by the first and add nothing, leaving $\Lambda_{s+f} = \Lambda_{s}$; the plotted value is therefore an upper bound, and the truth lies between it and $\Lambda_{s}$. Every point in that range leaves the self-checker short, so the approximation never favours our conclusion. It cannot justify overturning any untied vote, whatever stage-3 rule reads it: the shortfall is in the evidence, not in the rule. By contrast, gpt-oss-120b reaches 47.5, 11.8 and 48.9 on the three benchmarks. The self-check transcripts show the solver looking at the right evidence and misreading it. In 83\% of self-check calls on disagreeing breaking pairs (41 calls, development split) the solver's verification text engages the changed condition, yet it ratifies the solution that misuses it. In GSM8K pair 4 it quotes ``\$4 for each time she mows the lawn after the first 4 times'' word for word and then marks the trap's $4x$, which pays for every mowing, as \textsc{matches}.

\paragraph{Which splits the criterion licenses.}
Eq.~\eqref{eq:threshold} sets a different bar for every vote split, and the
32.4 above is the GSM8K bar for the strongest split four samples allow, not the
bar ESV usually faces. Re-estimating $\rho(v_m)/\rho(v_c)$ on each benchmark's
own candidates and counting overturns by split, 52 of
ESV's 66 GSM8K overturns, 67 of 72 on MATH-500 and 23 of 31 on JEEBench fall on
2--1--1, 2--2 or four-way splits, where the bar lies between 0.2 and 6.3 and
the measured $\Lambda_{\textsc{s}}$ of 47.5, 11.8 and 48.9 clears it with room
to spare. The MATH-500 result is licensed almost in full on this reading: the
five overturns of a 3--1 majority, worth 0.5 of the 5.9 points, are the only
ones near their bar, which is 27.9 there against $\Lambda_{\textsc{s}} = 11.8$
and an upper bound $\Lambda_{\textsc{s}+\textsc{f}} = 28.5$, and they returned
five fixes and no breaks. GSM8K behaves the same way: its 3--1 bar of 51.9 sits
just above $\Lambda_{\textsc{s}} = 47.5$, and those 14 overturns give 10 fixes
against 2 breaks, the only split on any benchmark where ESV breaks more than
one instance. Judge-select overturns strong majorities more readily. On
MATH-500 it takes 11 of the 3--1 splits for 8 fixes against 3 breaks, against
ESV's 5 and 0, which is one place its extra accuracy and its extra breaks both
come from.

\paragraph{Why the structured form loses.}
ESV abstains whenever no candidate is strictly sound (Eq.~\ref{eq:strict}), and abstentions grow from 72 instances on GSM8K to 379 on MATH-500 and 612 on JEEBench. Of the fixable instances (majority wrong, some candidate right), ESV abstains on 8 of 71, 51 of 121 and 109 of 138; on MATH-500 that is 50 abstentions plus one unreadable reply. Its false-positive rate stays at or below 5.2\%, but the share of correct candidates it passes falls from 86.7 to 60.8 to 25.0\%, and the ablation above rules out our decision rule, so the collapse is in the verdict itself. The obvious explanation, that longer derivations offer more to object to, does not hold either: the probability that a \textsc{flawed} candidate is longer than a \textsc{sound} one is 0.58 on GSM8K and 0.59 on MATH-500 but 0.42 on JEEBench, reversing on the benchmark where the collapse is worst. We therefore report the collapse as a limitation of contrastive verification rather than a diagnosis we have established. Judge-select also abstains often, on 68.4\% of disagreeing JEEBench instances, but a correct candidate existed in only 6.1\% of those, and the candidates it does name are right 98.1\% of the time.

\paragraph{What the perturbations show.}
Breaking pairs cost the majority vote 11.5 points against originals on GSM8K and 13.7 on MATH-500, while preserving pairs score above originals on both, so the difficulty comes from strategy misfit rather than novelty. On GSM8K, ESV gains most on breaking pairs (+8.4 points, against +3.8 on originals and +4.0 on preserving pairs). The checker helped accept every perturbed problem, so the honest test of a selection effect is the originals arm, which it never saw during construction. There ESV gains $+3.8$ (23/1), $+6.7$ (33/0) and $+2.8$ (11/0) over maj@4 and judge-select gains $+4.5$, $+9.8$ and $+13.2$, each significant, while USC is significant on none. Every ordering in Table~\ref{tab:main} survives on the unfiltered half, so the construction overlap does not drive the result. The trap explains 18.5\% of wrong majority answers on GSM8K but only 5.6\% on MATH-500, so we confine the strategy-misfit reading to GSM8K. There, ESV's labels recover the trap's known cause above chance: 75.6\% of trap candidates are labeled \textsc{binding} errors against 55.7\% of other wrong candidates (odds ratio 2.46, Fisher exact $p = 0.014$). The difference is not significant on MATH-500 ($p = 0.079$) and absent on JEEBench, and this interpretability on GSM8K is all the form adds.

\paragraph{Cost.}
ESV adds 43.6, 82.9 and 100.3\% tokens over maj@4 and judge-select 20.7, 48.7 and 68.5\%; per net fix ESV spends 2.4, 2.8 and 5.3 times as many. The gate's saving shrinks with difficulty: it skips 74.1\% of GSM8K instances but 10.1\% on JEEBench, where 46.2\% of the unanimous answers it passes unverified are wrong.
\section{Discussion and Limitations}
\paragraph{What the results support.}
A checker stronger than the solver can choose among the solver's own samples without a reference answer. The solver checking itself cannot: its \textsc{sound} verdict carries a likelihood ratio below 2, too weak for any decision rule to act on, and Eq.~\eqref{eq:threshold} says so before the verdict is measured. What the results do not support is our own verification format. Judge-select matches or beats ESV on every benchmark with the same checker and candidates at lower token cost, so for practitioners the recommendation is free-form judging by a stronger model, and the structured form's contribution is diagnostic rather than operational: it yields error labels recovering the trap's known cause on GSM8K, which free-form adjudication does not produce. Its strict rule also suits settings where overturning a correct majority costs more than missing a fix, since it breaks 2, 1 and 1 correct majorities against judge-select's 7, 5 and 0.

\paragraph{Why the form loses.}
The deficit is abstention, and the abstention is the checker's rather than the rule's (Table~\ref{tab:ablation}): demanding that every binding be quoted and every condition discharged is a bar the checker clears less and less often as problems harden, and we could not establish why. A fix would therefore have to change what the checker is asked to certify, not how its answer is parsed.

\paragraph{Limitations.}
The checker differs from the solvers in family, total size (116.8B against 4--12B) and reasoning training, and we cannot tell which difference matters. It also helped accept the perturbed problems, so those arms measure it on a set its own competence filtered; the originals arm, which no filter touched, reproduces every ordering, which argues against a selection effect but does not exclude one. Two controls are missing. The first is four extra samples at matched compute (maj@8): subsampling the stored samples shows the fourth already worth only $+0.6$, $+1.4$ and $+0.9$ points on a decelerating curve, so doubling the budget would have to beat a per-sample return under 1.5 points to match a 5-point gain, which we think unlikely but have not shown. The second is the checker answering each problem directly. On perturbed problems that is uninformative as well as unrun, since acceptance required the checker to solve $P'$ correctly while blind to $P$, putting it at ceiling by construction; on originals it is genuinely unmeasured, and if it beat every selector it would narrow the practical case for selection, though not the comparison among selectors, which share one solver pool. We draw one sample set per instance, so seed variance is unmeasured; this bears on the 0.6-point GSM8K gap between judge-select and ESV, which rests on 7 paired disagreements, and much less on gaps resting on tens of them. Our exact tests treat instances as independent; a cluster bootstrap over source problems leaves every conclusion standing. Judge-select was added after the held-out run, JEEBench shaped an earlier pipeline and is marked as development throughout, and all results are on mathematics with small solvers. Pair validation rests on agreement between two verifier models with a 19-pair hand audit on GSM8K alone; a stratified human audit across all three benchmarks is the check we owe this benchmark.

\section{Conclusion and Future Work}
We asked whether a model's sampled answers can be filtered without a reference answer, without re-solving, and without breaking correct majorities. They can, but the component that decides the outcome is not the one we set out to test. On a paired perturbation benchmark, a stronger checker raises accuracy over majority voting by 3.5 to 12.7 points while the solver checking itself lowers it on every benchmark, and a likelihood-ratio criterion fixed on development data predicts both outcomes in advance. Our structured verification format, by contrast, is beaten by free-form adjudication with the same checker on all three benchmarks; its abstention traces to the checker's verdict rather than to our decision rule, and what it adds is interpretable error labels on GSM8K rather than accuracy. We therefore report the format as a negative result and the checker-strength criterion as the finding worth carrying forward. Next we will run the matched-compute and checker-alone baselines, replicate across seeds, vary checker size within one model family to separate strength from independence, audit the pairs with human annotators, and test the criterion beyond mathematics.
\bibliography{aaai2027}   

@inproceedings{wei2022chain,
  title     = {Chain-of-Thought Prompting Elicits Reasoning in Large Language Models},
  author    = {Wei, Jason and Wang, Xuezhi and Schuurmans, Dale and Bosma, Maarten
               and Ichter, Brian and Xia, Fei and Chi, Ed and Le, Quoc V. and Zhou, Denny},
  booktitle = {Advances in Neural Information Processing Systems (NeurIPS)},
  volume    = {35},
  pages     = {24824--24837},
  year      = {2022}
}

@inproceedings{wang2022self,
  title     = {Self-Consistency Improves Chain of Thought Reasoning in Language Models},
  author    = {Wang, Xuezhi and Wei, Jason and Schuurmans, Dale and Le, Quoc and Chi, Ed
               and Narang, Sharan and Chowdhery, Aakanksha and Zhou, Denny},
  booktitle = {International Conference on Learning Representations (ICLR)},
  year      = {2023}
}

@inproceedings{madaan2023self,
  title     = {Self-Refine: Iterative Refinement with Self-Feedback},
  author    = {Madaan, Aman and Tandon, Niket and Gupta, Prakhar and Hallinan, Skyler
               and Gao, Luyu and Wiegreffe, Sarah and Alon, Uri and Dziri, Nouha
               and Prabhumoye, Shrimai and Yang, Yiming and others},
  booktitle = {Advances in Neural Information Processing Systems (NeurIPS)},
  volume    = {36},
  year      = {2023}
}

@inproceedings{shinn2023reflexion,
  title     = {Reflexion: Language Agents with Verbal Reinforcement Learning},
  author    = {Shinn, Noah and Cassano, Federico and Gopinath, Ashwin
               and Narasimhan, Karthik and Yao, Shunyu},
  booktitle = {Advances in Neural Information Processing Systems (NeurIPS)},
  volume    = {36},
  year      = {2023}
}

@article{chen2023universal,
  title   = {Universal Self-Consistency for Large Language Model Generation},
  author  = {Chen, Xinyun and Aksitov, Renat and Alon, Uri and Ren, Jie and Xiao, Kefan
            and Yin, Pengcheng and Prakash, Sushant and Sutton, Charles
            and Wang, Xuezhi and Zhou, Denny},
  journal = {arXiv preprint arXiv:2311.17311},
  year    = {2023}
}

@inproceedings{huang2024large,
  title     = {Large Language Models Cannot Self-Correct Reasoning Yet},
  author    = {Huang, Jie and Chen, Xinyun and Mishra, Swaroop and Zheng, Huaixiu Steven
               and Yu, Adams Wei and Song, Xinying and Zhou, Denny},
  booktitle = {International Conference on Learning Representations (ICLR)},
  year      = {2024}
}

@article{cobbe2021training,
  title   = {Training Verifiers to Solve Math Word Problems},
  author  = {Cobbe, Karl and Kosaraju, Vineet and Bavarian, Mohammad and Chen, Mark
            and Jun, Heewoo and Kaiser, Lukasz and Plappert, Matthias and Tworek, Jerry
            and Hilton, Jacob and Nakano, Reiichiro and Hesse, Christopher and Schulman, John},
  journal = {arXiv preprint arXiv:2110.14168},
  year    = {2021}
}

@inproceedings{lightman2024let,
  title     = {Let's Verify Step by Step},
  author    = {Lightman, Hunter and Kosaraju, Vineet and Burda, Yura and Edwards, Harrison
               and Baker, Bowen and Lee, Teddy and Leike, Jan and Schulman, John
               and Sutskever, Ilya and Cobbe, Karl},
  booktitle = {International Conference on Learning Representations (ICLR)},
  year      = {2024}
}

@inproceedings{zheng2023judging,
  title     = {Judging LLM-as-a-Judge with MT-Bench and Chatbot Arena},
  author    = {Zheng, Lianmin and Chiang, Wei-Lin and Sheng, Ying and Zhuang, Siyuan
               and Wu, Zhanghao and Zhuang, Yonghao and Lin, Zi and Li, Zhuohan
               and Li, Dacheng and Xing, Eric P. and others},
  booktitle = {Advances in Neural Information Processing Systems (NeurIPS):
               Datasets and Benchmarks Track},
  volume    = {36},
  year      = {2023}
}

@inproceedings{mirzadeh2025gsm,
  title     = {{GSM-Symbolic}: Understanding the Limitations of Mathematical Reasoning
               in Large Language Models},
  author    = {Mirzadeh, Iman and Alizadeh, Keivan and Shahrokhi, Hooman and Tuzel, Oncel
               and Bengio, Samy and Farajtabar, Mehrdad},
  booktitle = {International Conference on Learning Representations (ICLR)},
  year      = {2025},
  note      = {arXiv:2410.05229}
}

@article{snell2024scaling,
  title   = {Scaling {LLM} Test-Time Compute Optimally can be More Effective than
            Scaling Model Parameters},
  author  = {Snell, Charlie and Lee, Jaehoon and Xu, Kelvin and Kumar, Aviral},
  journal = {arXiv preprint arXiv:2408.03314},
  year    = {2024}
}

@inproceedings{hendrycks2021math,
  title     = {Measuring Mathematical Problem Solving With the {MATH} Dataset},
  author    = {Hendrycks, Dan and Burns, Collin and Kadavath, Saurav and Arora, Akul
               and Basart, Steven and Tang, Eric and Song, Dawn and Steinhardt, Jacob},
  booktitle = {Advances in Neural Information Processing Systems (NeurIPS):
               Datasets and Benchmarks Track},
  year      = {2021}
}

@inproceedings{arora2023jeebench,
  title     = {Have {LLMs} Advanced Enough? A Challenging Problem Solving Benchmark
               for Large Language Models},
  author    = {Arora, Daman and Singh, Himanshu Gaurav and Mausam},
  booktitle = {Proceedings of the 2023 Conference on Empirical Methods in Natural
               Language Processing (EMNLP)},
  pages     = {7527--7543},
  year      = {2023}
}

@article{qwen2025qwen25,
  title   = {{Qwen2.5} Technical Report},
  author  = {{Qwen Team}},
  journal = {arXiv preprint arXiv:2412.15115},
  year    = {2025}
}

@article{gemma2025gemma3,
  title   = {{Gemma 3} Technical Report},
  author  = {{Gemma Team}},
  journal = {arXiv preprint arXiv:2503.19786},
  year    = {2025}
}

@article{openai2025gptoss,
  title   = {{gpt-oss-120b} and {gpt-oss-20b} Model Card},
  author  = {{OpenAI}},
  journal = {arXiv preprint arXiv:2508.10925},
  year    = {2025}
}

@article{mcnemar1947,
  title   = {Note on the Sampling Error of the Difference between Correlated
            Proportions or Percentages},
  author  = {McNemar, Quinn},
  journal = {Psychometrika},
  volume  = {12},
  number  = {2},
  pages   = {153--157},
  year    = {1947}
}

\end{document}


\appendix
\renewcommand{\topfraction}{0.95}
\renewcommand{\bottomfraction}{0.9}
\renewcommand{\textfraction}{0.05}
\setcounter{topnumber}{4}
\setcounter{bottomnumber}{3}
\setcounter{totalnumber}{8}
\section{Appendix}

\subsection{Code and data}
\label{app:code}

The frozen perturbation sets, every raw model response, the decision-rule
implementation and the analysis scripts that regenerate each table and figure in
this paper are available at:

\begin{center}
\small\url{https://anonymous.4open.science/r/Project_structural-perturbations-esv-09E7/}
\end{center}

\noindent
Everything is released under the Apache License 2.0. The inventory is: the three
\texttt{frozen\_dataset.json} files of $(P, P')$ pairs with gold answers, named
strategy, changed condition and per-verifier validation; the
\texttt{generation\_log.jsonl} for each benchmark, recording every perturbation
attempt including the rejected ones with their rejection reason; the per-solver
\texttt{records.json} holding all four Direct QA and four CoT samples per
instance with their raw text, token usage and latency; the ESV, USC and
judge-select verification records under each checker, with every raw checker
reply and its parsed fields; the full prompt sources
(\texttt{method\_prompts.py}, \texttt{perturb\_prompts.py}), the decision-rule
and scoring implementation (\texttt{run\_esv3.py}, \texttt{common.py},
\texttt{expression\_answers.py}), and the analyzer
(\texttt{analyze\_esv3.py}) that regenerates every table and figure in this
paper. The repository is roughly 140\,MB, dominated by the raw responses, which
are kept so that each number is auditable to the call that produced it. No model
call is needed to reproduce a table: the stored responses are sufficient, and
the analyzer runs offline. We commit to a permanent, de-anonymized public
release of the same artefacts under the same license regardless of the outcome
of review; the de-anonymized link will replace the one above in the
camera-ready version.

\subsection{Datasets}
\label{app:data}

\begin{table}[!htbp]
\centering
\small
\setlength{\tabcolsep}{3pt}
\begin{tabular}{@{}lrrr@{}}
\toprule
 & GSM8K & MATH-500 & JEEBench \\
\midrule
Usable source problems & 1{,}317 & 500 & 236 \\
Sampled (seed 2026) & 200 & 250 & 200 \\
Frozen pairs & 198 & 244 & 193 \\
\quad breaking/preserving & 99/99 & 123/121 & 97/96 \\
Acceptance, brk./pres. & 1.00/0.98 & 0.98/0.97 & 0.95/0.98 \\
Attempts (1/2/3+) & 182/16/0 & 216/23/5 & 155/25/13 \\
Solvers & 3 & 2 & 2 \\
Scored instances & 1{,}170 & 976 & 772 \\
Disagreeing & 25.9\% & 61.6\% & 89.9\% \\
Headroom (points) & 6.1 & 12.4 & 17.9 \\
\bottomrule
\end{tabular}
\caption{The frozen perturbation sets. Instances count $P$ and $P'$ once per
solver. Acceptance is the share of attempted perturbations the two verifiers
accepted, per mode; attempts counts how many generations a frozen pair needed.
GSM8K drops two source problems with negative answers and excludes six
originals with disputed gold answers, leaving 390 instances per solver.}
\label{tab:datasets}
\end{table}

Table~\ref{tab:datasets} summarizes the three frozen sets.
GSM8K answers are all integers. MATH-500 keeps 145 integer, 88 expression and 11 numeric answers, and its 244 pairs span levels 1--5 (20/43/58/61/62) and all seven subjects (Algebra 59, Intermediate Algebra 47, Prealgebra 44, Precalculus 30, Number Theory 27, Geometry 21, Counting and Probability 16). JEEBench mathematics answers are multiple-correct (68), numeric (55), single-correct (41) and integer (29). Expression answers are compared by a normalizer that tries canonical and case-insensitive string equality, element-wise comparison of tuples, intervals, sets and vectors with matching delimiters, exact numeric equality, and finally a symbolic check that is abandoned unless the \LaTeX{} conversion is complete; a decimal approximation of an exact value is never accepted. A hand audit of 19 GSM8K pairs found every accepted answer correct. The six excluded GSM8K originals are pairs 40, 82, 95, 105, 111 and 189; their perturbed siblings keep their verified answers. On GSM8K the held-out split (pairs 60--197) gives maj@4 88.8\%, USC 89.7\%, self-check ESV 85.1\%, ESV with gpt-oss-120b 94.1\% and pass@4 95.1\%; the development split gives 87.7, 88.5, 85.2, 92.2 and 93.3\%. The judge-select control was designed and run after the held-out evaluation and was not part of the pre-registered criterion.

\subsection{Exact tests and confidence intervals}
\label{app:pvalues}

Table~\ref{tab:pvalues} gives, for every method, the accuracy change against CoT maj@4 and the exact two-sided McNemar $p$-value on the discordant instances. Table~\ref{tab:h2h} gives the head-to-head comparisons quoted in the text.

\paragraph{Clustering.} The McNemar tests treat instances as independent, which
they are not: one source problem yields $P$ and $P'$ and is scored under two or
three solvers, so it contributes up to six correlated instances. We therefore
repeated every comparison with a nonparametric bootstrap that resamples whole
source problems rather than instances, 5{,}000 resamples at the level of the
198, 244 and 193 frozen pairs, which absorbs both the $P$/$P'$ pairing and the
sharing across solvers. Every conclusion survives and no interval changes sign.
The 95\% clustered intervals for \method{} are $[+3.6, +6.7]$, $[+4.5, +7.4]$
and $[+2.2, +4.9]$ points and for judge-select $[+4.1, +7.3]$, $[+7.7, +11.5]$
and $[+10.4, +15.2]$, so both clear zero everywhere; self-check stays entirely
below zero on GSM8K and MATH-500; and USC covers zero on all three, agreeing
with the exact tests that it does not beat the vote. We treat the exact tests as
primary, since they were fixed before the held-out run, and the bootstrap as a
secondary check on the independence assumption. Both regenerate from the stored
responses with the released analyzer.

\begin{table*}[t]
\centering
\small
\setlength{\tabcolsep}{3pt}
\begin{tabular}{lrrrrrr}
\toprule
 & \multicolumn{2}{c}{GSM8K} & \multicolumn{2}{c}{MATH-500} & \multicolumn{2}{c}{JEEBench} \\
\cmidrule(lr){2-3}\cmidrule(lr){4-5}\cmidrule(lr){6-7}
Method & $\Delta$ & $p$ & $\Delta$ & $p$ & $\Delta$ & $p$ \\
\midrule
Direct QA & $-76.3$ & 2e-256 & $-45.7$ & 6e-110 & $-12.0$ & 8e-11 \\
USC & $+0.9$ & 0.087 & $+0.4$ & 0.71 & $-1.9$ & 0.11 \\
ESV, self & $-3.3$ & 7e-06 & $-2.3$ & 0.017 & $-2.2$ & 0.078 \\
ESV, 120b & $+5.0$ & 4e-16 & $+5.9$ & 1e-16 & $+3.5$ & 1e-07 \\
Judge-select & $+5.6$ & 5e-19 & $+9.6$ & 1e-23 & $+12.7$ & 6e-30 \\
pass@4 & $+6.1$ & 8e-22 & $+12.4$ & 8e-37 & $+17.9$ & 6e-42 \\
\bottomrule
\end{tabular}
\caption{Change in accuracy over CoT maj@4 (points) and exact two-sided McNemar $p$, paired on the same instances.}
\label{tab:pvalues}
\end{table*}

\begin{table}[!htbp]
\centering
\small
\setlength{\tabcolsep}{4pt}
\begin{tabular}{@{}lccc@{}}
\toprule
Comparison & GSM8K & MATH-500 & JEEBench \\
\midrule
\method{} vs.\ USC   & 53 / 4 & 71 / 17 & 69 / 27 \\
                     & 6e-12 & 5e-09 & 2e-05 \\
\method{} vs.\ Judge & 0 / 7 & 4 / 40 & 2 / 73 \\
                     & 0.016 & 2e-08 & 2e-19 \\
Cascade vs.\ Judge   & 0 / 2 & 2 / 3 & 2 / 2 \\
                     & 0.50 & 1.00 & 1.00 \\
Self-check vs.\ USC  & 17 / 66 & 35 / 61 & 45 / 47 \\
                     & 6e-08 & 0.010 & 0.92 \\
\bottomrule
\end{tabular}
\caption{Paired head-to-head comparisons on the same instances, with \method{}
using the gpt-oss-120b checker unless stated and Judge the judge-select control.
The first line of each block gives wins / losses for the left-hand method, the
second the exact two-sided McNemar $p$. The cascade runs \method{} and falls
back to judge-select when no candidate passes the strict rule.}
\label{tab:h2h}
\end{table}

\subsection{Decisions, selection quality and checker rates}
\label{app:decisions}

Table~\ref{tab:decisions} breaks ESV's outcomes down by what the decision rule did. The last column counts instances where the majority was wrong, a candidate was right, and the rule abstained; these are the gains a looser rule could still reach. Table~\ref{tab:selection} restricts accuracy to disagreeing instances, the only ones a selector can change, and Table~\ref{tab:detector} gives the per-candidate checker rates behind Figure~\ref{fig:lr}.

\begin{table}[!htbp]
\centering
\small
\setlength{\tabcolsep}{3pt}
\begin{tabular}{@{}lrrrrr@{}}
\toprule
Rule outcome & $n$ & maj@4 & \method{} & $F$ / $B$ & Decl. \\
\midrule
\multicolumn{6}{@{}l}{\textit{GSM8K} (1{,}170)} \\
\quad unanimous      & 867 & 98.0 & 98.0 & --     & -- \\
\quad one sound      & 231 & 70.6 & 96.1 & 61 / 2 & -- \\
\quad none (abstain) &  72 & 30.6 & 30.6 & --     & 8 \\
\midrule
\multicolumn{6}{@{}l}{\textit{MATH-500} (976)} \\
\quad unanimous      & 375 & 96.0 & 96.0 & --     & -- \\
\quad one sound      & 187 & 59.4 & 89.3 & 57 / 1 & -- \\
\quad several sound  &  34 & 38.2 & 44.1 & 2 / 0  & -- \\
\quad none (abstain) & 379 & 17.4 & 17.4 & --     & 50 \\
\quad unreadable     &   1 &  0.0 &  0.0 & --     & 1 \\
\midrule
\multicolumn{6}{@{}l}{\textit{JEEBench} (772)} \\
\quad unanimous      &  78 & 53.8 & 53.8 & --     & -- \\
\quad one sound      &  75 & 56.0 & 92.0 & 28 / 1 & -- \\
\quad several sound  &   2 &  0.0 &  0.0 & 0 / 0  & -- \\
\quad none (abstain) & 612 & 16.5 & 16.5 & --     & 109 \\
\quad unreadable     &   5 &  0.0 &  0.0 & --     & -- \\
\bottomrule
\end{tabular}
\caption{\method{} (gpt-oss-120b) by decision-rule outcome; accuracies in \%.
``one sound'' is the branch that commits to a candidate, ``Decl.'' counts
fixable instances (majority wrong, a candidate right) on which the rule
abstained. GSM8K has no \emph{several sound} or \emph{unreadable} instances, so
those rows are absent rather than zero. The rows sum to the scored instances of
each benchmark.}
\label{tab:decisions}
\end{table}

\begin{table}[!htbp]
\centering
\small
\setlength{\tabcolsep}{4pt}
\begin{tabular}{@{}lrrr@{}}
\toprule
Selector & GSM8K & MATH & JEE \\
$n$ disagreeing & 303 & 601 & 694 \\
\midrule
Random candidate & 37.0 & 20.0 & 15.6 \\
Majority vote & 61.1 & 31.6 & 20.6 \\
USC & 64.4 & 32.3 & 18.4 \\
\method{}, self-check & 48.2 & 28.0 & 18.2 \\
\method{}, gpt-oss-120b & 80.5 & 41.3 & 24.5 \\
Judge-select & \textbf{82.8} & \textbf{47.3} & \textbf{34.7} \\
Oracle (pass@4) & 84.5 & 51.7 & 40.5 \\
\bottomrule
\end{tabular}
\caption{Selection quality: accuracy (\%) on instances whose four samples disagree. The random-candidate row is the expected accuracy of drawing one candidate uniformly, averaged over instances.}
\label{tab:selection}
\end{table}

\begin{table}[!htbp]
\centering
\small
\setlength{\tabcolsep}{3pt}
\begin{tabular}{@{}llrrrrr@{}}
\toprule
 & Check & TPR & FPR & Prec. & $\Lambda_{\textsc{s}}$ & $\Lambda_{\textsc{s}+\textsc{f}}$ \\
\midrule
GSM8K & self & 57.0 & 36.5 & 44.8 & 1.6 & 2.3 \\
      & 120b & 86.7 & 1.8 & 96.1 & 47.5 & 351 \\
MATH  & self & 56.3 & 43.3 & 20.7 & 1.3 & 1.7 \\
      & 120b & 60.8 & 5.2 & 70.3 & 11.8 & 28.5 \\
JEE   & self & 44.6 & 42.5 & 14.3 & 1.1 & 1.1 \\
      & 120b & 25.0 & 0.5 & 88.6 & 48.9 & 64.9 \\
\bottomrule
\end{tabular}
\caption{Per-candidate checker rates; ``120b'' is gpt-oss-120b. TPR and FPR
are the shares of correct and of wrong candidates judged strictly
\textsc{sound}, Prec.\ the precision of that verdict.
$\Lambda_{\textsc{s}}$ is its likelihood ratio and
$\Lambda_{\textsc{s}+\textsc{f}}$ multiplies in \textsc{flawed} on a competing
candidate, assuming independence. Candidates: 256/493, 311/1{,}552 and
280/1{,}760 correct/wrong.}
\label{tab:detector}
\end{table}

\begin{table}[!htbp]
\centering
\small
\begin{tabular}{lrrrr}
\toprule
Split & $n$ & Eq.~\eqref{eq:threshold} bar & Overturns & F / B \\
\midrule
\multicolumn{5}{l}{\emph{GSM8K}, $\Lambda_{\textsc{s}} = 47.5$} \\
3--1 & 158 & 51.9 & 14 & 10 / 2 \\
2--1--1 & 81 & 6.3 & 25 & 24 / 0 \\
2--2 & 33 & 0.2 & 18 & 18 / 0 \\
1--1--1--1 & 31 & 0.7 & 9 & 9 / 0 \\
\midrule
\multicolumn{5}{l}{\emph{MATH-500}, $\Lambda_{\textsc{s}} = 11.8$} \\
3--1 & 126 & 27.9 & 5 & 5 / 0 \\
2--1--1 & 184 & 3.4 & 34 & 27 / 1 \\
2--2 & 47 & 2.2 & 5 & 5 / 0 \\
1--1--1--1 & 235 & 1.2 & 28 & 22 / 0 \\
other & 9 & --- & 0 & 0 / 0 \\
\midrule
\multicolumn{5}{l}{\emph{JEEBench}$^{\ddagger}$, $\Lambda_{\textsc{s}} = 48.9$} \\
3--1 & 142 & 3.6 & 8 & 7 / 1 \\
2--1--1 & 229 & 2.5 & 8 & 8 / 0 \\
2--2 & 47 & 0.6 & 4 & 4 / 0 \\
1--1--1--1 & 191 & 1.0 & 8 & 6 / 0 \\
other & 85 & 1.5 & 3 & 3 / 0 \\
\bottomrule
\end{tabular}
\caption{ESV overturns by the vote split it overturned, pooled over solvers.
The bar is $\rho(v_m)/\rho(v_c)$ from Eq.~\eqref{eq:threshold}, re-estimated on
that benchmark's own candidates and conditioned on the whole split rather than
on the majority's vote count alone, so it is comparable with the $\Lambda$
measured on the same benchmark. $n$ counts disagreeing instances with that
split; an overturn is a committed change away from the majority answer; F / B
are fixes and breaks against CoT maj@4, and they sum per benchmark to the
61/2, 59/1 and 28/1 of Table~\ref{tab:main}. \emph{other} pools splits whose
four samples did not all parse.}
\label{tab:splits}
\end{table}

When judge-select names a candidate, that candidate is correct 96.5\% (GSM8K), 89.6\% (MATH-500) and 98.1\% (JEEBench) of the time; it answers NONE on 15.2\%, 49.8\% and 68.4\% of disagreeing instances.

\begin{table}[!htbp]
\centering
\footnotesize
\begin{tabular}{lrcrcrc}
\toprule
 & \multicolumn{2}{c}{GSM8K (576)} & \multicolumn{2}{c}{MATH (488)} & \multicolumn{2}{c}{JEE$^{\ddagger}$ (386)} \\
\cmidrule(lr){2-3} \cmidrule(lr){4-5} \cmidrule(lr){6-7}
Method & Acc & $F$/$B$ & Acc & $F$/$B$ & Acc & $F$/$B$ \\
\midrule
CoT maj@4        & 90.6 & ---   & 57.6 & ---   & 23.6 & --- \\
USC              & 91.5 & 9/4   & 58.6 & 15/10 & 21.5 & 14/22 \\
\method{}        & 94.4 & 23/1  & 64.3 & 33/0  & 26.4 & 11/0 \\
Judge-select     & \textbf{95.1} & 26/0  & \textbf{67.4} & 51/3  & \textbf{36.8} & 51/0 \\
\midrule
CoT pass@4       & 95.7 & ---   & 70.1 & ---   & 41.5 & --- \\
\bottomrule
\end{tabular}
\caption{Every selector on the untouched originals only. gpt-oss-120b was one
of the two models that accepted each perturbed problem, so the perturbed arms
of Table~\ref{tab:main} measure the checker on items its own competence
filtered; the originals passed through no such filter. $F$/$B$ are fixes and
breaks against CoT maj@4 on the same instances. \method{} beats maj@4 on all
three ($p = 3\times10^{-6}$, $2\times10^{-10}$, $1\times10^{-3}$) and so does
judge-select ($3\times10^{-8}$, $3\times10^{-12}$, $9\times10^{-16}$), while
USC reaches significance on none ($p = 0.27$, $0.42$, $0.24$). The ordering of
Table~\ref{tab:main} is unchanged.}
\label{tab:originals}
\end{table}

\paragraph{Choices JEEBench shaped.}
Four decisions were made or discarded while looking at JEEBench results, which
is why Section~\ref{sec:setup} treats it as development rather than held-out
evidence: the move from a five-stage pipeline with diagnosis-driven repair to
the three stages reported here, the disagreement gate, the strict form of
Eq.~\eqref{eq:strict}, and dropping an oracle-triggered variant that used gold
to decide when to verify.

\subsection{Perturbation arms and answer formats}
\label{app:arms}

Table~\ref{tab:arms} splits every instance by which arm of the pair it came
from: the untouched original $P$, a strategy-breaking $P'$, or a
strategy-preserving $P'$. The breaking arm is the one \method{} was designed
for, so the fixes-and-breaks column is read per arm.

\begin{table}[t]
\centering
\small
\setlength{\tabcolsep}{3.5pt}
\begin{tabular}{@{}llrrrr@{}}
\toprule
 & Arm & $n$ & maj@4 & \method{} & $F$ / $B$ \\
\midrule
GSM8K & originals & 576 & 90.6 & 94.4 & 23 / 1 \\
      & breaking $P'$ & 297 & 79.1 & 87.5 & 25 / 0 \\
      & preserv.\ $P'$ & 297 & 93.6 & 97.6 & 13 / 1 \\
\midrule
MATH  & originals & 488 & 57.6 & 64.3 & 33 / 0 \\
      & breaking $P'$ & 246 & 43.9 & 50.0 & 16 / 1 \\
      & preserv.\ $P'$ & 242 & 66.5 & 70.7 & 10 / 0 \\
\midrule
JEE   & originals & 386 & 23.6 & 26.4 & 11 / 0 \\
      & breaking $P'$ & 194 & 24.7 & 27.8 & 7 / 1 \\
      & preserv.\ $P'$ & 192 & 24.0 & 29.2 & 10 / 0 \\
\bottomrule
\end{tabular}
\caption{Accuracy (\%) by perturbation arm, ESV with gpt-oss-120b. USC on GSM8K scores 91.5, 79.8 and 94.6 on the three arms. Only on GSM8K and MATH-500 does the breaking arm sit below the originals; on JEEBench the three arms are within 1.1 points of each other, which is why we confine the strategy-misfit reading to the first two.}
\label{tab:arms}
\end{table}

On GSM8K the trap answer accounts for 25 of the 135 wrong majority answers (18.5\%) and appears in 90 of the 1{,}188 samples drawn on breaking pairs (7.6\%); 54 of the wrong answers (40.0\%) are on unperturbed originals. On MATH-500 the trap accounts for 24 of 426 wrong answers (5.6\%).

MATH-500 answers come in three formats. ESV's gain is similar across them: expression answers (352 instances) rise from 40.6 to 46.0\% (20/1), integers (580) from 65.5 to 71.9\% (37/0) and numeric answers (44) from 61.4 to 65.9\% (2/0). The expression scorer is therefore not what separates the methods.

\subsection{Error attribution}
\label{app:attribution}

On a breaking pair, a candidate equal to the trap applies the original strategy as if the changed condition still held, so its cause is known. Table~\ref{tab:attribution} compares ESV's failure-type labels on these candidates with its labels on other wrong candidates. The labels are informative but not specific: 37--61\% of all wrong candidates are labeled \textsc{binding}.

\begin{table*}[!htbp]
\centering
\small
\setlength{\tabcolsep}{3pt}
\begin{tabular}{llrrrrr}
\toprule
 & Candidate group & $n$ & \textsc{bind.} & \textsc{meth.} & \textsc{comp.} & \textsc{sound} \\
\midrule
GSM8K & trap (known cause) & 45 & 75.6 & 17.8 & 6.7 & 0.0 \\
 & other wrong, breaking & 193 & 60.6 & 15.0 & 24.4 & 0.0 \\
 & wrong, orig.\ or pres. & 255 & 55.7 & 9.8 & 31.0 & 3.5 \\
 & correct & 256 & 9.8 & 1.6 & 2.0 & 86.7 \\
\midrule
MATH-500 & trap & 43 & 51.2 & 27.9 & 20.9 & 0.0 \\
 & wrong, orig.\ or pres. & 1{,}062 & 37.5 & 24.9 & 31.2 & 6.2 \\
\midrule
JEEBench & trap & 57 & 42.1 & 31.6 & 21.1 & 3.5 \\
 & wrong, orig.\ or pres. & 1{,}320 & 46.9 & 29.5 & 22.7 & 0.3 \\
\bottomrule
\end{tabular}
\caption{ESV's failure-type labels (\% of candidates, gpt-oss-120b checker) by ground-truth group. A trap candidate is a \textsc{binding} error by construction. Trap against other wrong candidates on originals and preserving pairs, \textsc{binding} rate: GSM8K OR 2.46 ($p$=0.014), MATH-500 OR 1.75 ($p$=0.079), JEEBench OR 0.82 ($p$=0.501), Fisher exact test.}
\label{tab:attribution}
\end{table*}

\subsection{Supporting figures and tables}
\label{app:support}

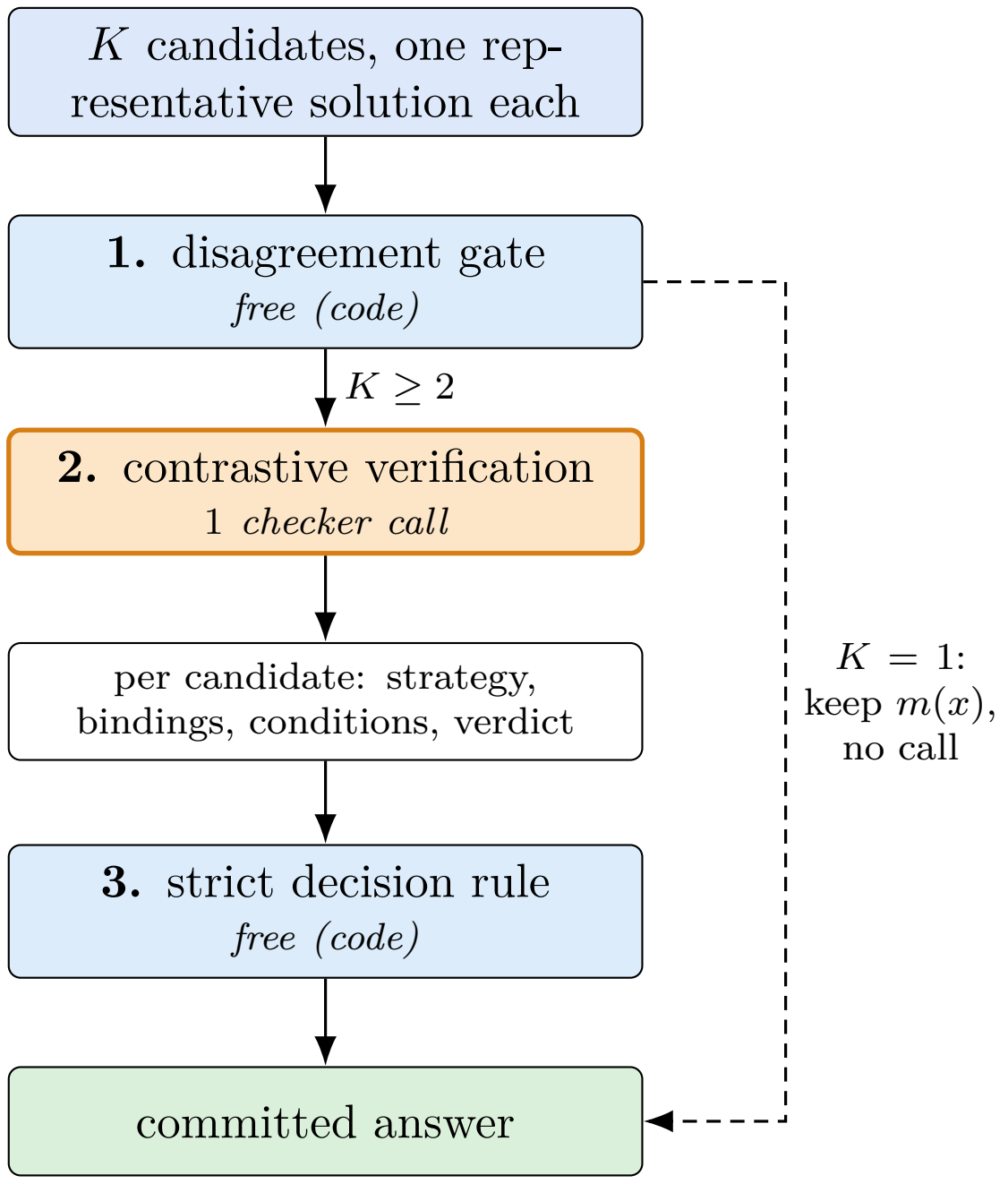
\begin{figure}[!htbp]
\centering
\begin{tikzpicture}[
  font=\small,
  base/.style={draw, rounded corners=2.5pt, align=center, inner sep=4pt,
               text width=44mm, minimum height=8mm},
  shared/.style={base, fill=cshared},
  code/.style={base, fill=ccode},
  call/.style={base, fill=ccall, draw=ccallb, line width=1pt},
  outbox/.style={base, fill=cout},
  fields/.style={base, fill=white, font=\scriptsize},
  ar/.style={-{Latex[length=2.2mm]}, semithick},
  cost/.style={font=\scriptsize},
  side/.style={font=\scriptsize, align=center, text width=15mm},
]
\node[shared] (cands) at (0,0)     {$K$ candidates, one representative solution each};
\node[code]   (gate)  at (0,-1.55) {\textbf{1.} disagreement gate\\[-1pt]{\cost free (code)}};
\node[call]   (ver)   at (0,-3.10) {\textbf{2.} contrastive verification\\[-1pt]{\cost $1$ checker call}};
\node[fields] (flds)  at (0,-4.65) {per candidate: strategy, bindings, conditions, verdict};
\node[code]   (rule)  at (0,-6.20) {\textbf{3.} strict decision rule\\[-1pt]{\cost free (code)}};
\node[outbox] (ans)   at (0,-7.75) {committed answer};

\draw[ar] (cands) -- (gate);
\draw[ar] (gate)  -- node[right, cost] {$K \ge 2$} (ver);
\draw[ar] (ver)   -- (flds);
\draw[ar] (flds)  -- (rule);
\draw[ar] (rule)  -- (ans);

\coordinate (b1) at (3.4,-1.55);
\coordinate (b2) at (3.4,-7.75);
\draw[ar, densely dashed] (gate.east) -- (b1) -- (b2) -- (ans.east);
\node[side] at (4.25,-4.65) {$K = 1$:\\keep $m(x)$,\\no call};
\end{tikzpicture}
\caption{Inside \method{}, shaded by what each step costs. Two of the three
stages are deterministic code and free (blue); only contrastive verification
issues a model call (orange, heavy border), and only on the $K \ge 2$ instances
where the samples disagree. Unanimous instances take the dashed path and cost
nothing beyond sampling. The checker returns typed per-candidate fields rather
than a choice, and stage~3 reads those fields in code, so the model supplies
evidence and the rule supplies the decision.}
\label{fig:pipeline}
\end{figure}

\begin{figure}[!htbp]
\centering
\begin{tikzpicture}[x=1cm, y=0.185cm, font=\footnotesize]
\fill[black!8] (0.70,0) rectangle (2.45,6.1);
\fill[black!8] (3.00,0) rectangle (4.75,12.4);
\fill[black!8] (5.30,0) rectangle (7.05,17.9);
\node[anchor=south, font=\scriptsize, black!50] at (1.575,6.2)  {$+6.1$};
\node[anchor=south, font=\scriptsize, black!50] at (3.875,12.5) {$+12.4$};
\node[anchor=south, font=\scriptsize, black!50] at (6.175,18.0) {$+17.9$};
\draw[black!30, densely dashed] (0.56,0) -- (7.20,0);
\draw[black!55] (0.56,-6) -- (0.56,19.2);
\foreach \y in {-5,0,5,10,15}
  \draw[black!55] (0.49,\y) -- (0.56,\y) node[left=0pt, font=\scriptsize] {$\y$};
\node[anchor=south, font=\scriptsize, align=center] at (0.10,13.0) {pts.\\vs.\\maj@4};
\node[font=\scriptsize] at (1.575,-9.2) {GSM8K};
\node[font=\scriptsize] at (3.875,-9.2) {MATH-500};
\node[font=\scriptsize] at (6.175,-9.2) {JEEBench};
\ebar{1.00}{0.9}{0.9}{pgrey}   \mkopensq{(1.00,0.9)}{pgrey}
\ebar{1.38}{-3.3}{1.4}{pgrey}  \mkopenci{(1.38,-3.3)}{pgrey}
\ebar{1.76}{5.0}{1.3}{ccallb}  \mkfillci{(1.76,5.0)}{ccallb}
\ebar{2.14}{5.6}{1.3}{pblue}   \mkfillsq{(2.14,5.6)}{pblue}
\ebar{3.30}{0.4}{1.6}{pgrey}   \mkopensq{(3.30,0.4)}{pgrey}
\ebar{3.68}{-2.3}{1.8}{pgrey}  \mkopenci{(3.68,-2.3)}{pgrey}
\ebar{4.06}{5.9}{1.5}{ccallb}  \mkfillci{(4.06,5.9)}{ccallb}
\ebar{4.44}{9.6}{2.0}{pblue}   \mkfillsq{(4.44,9.6)}{pblue}
\ebar{5.60}{-1.9}{2.3}{pgrey}  \mkopensq{(5.60,-1.9)}{pgrey}
\ebar{5.98}{-2.2}{2.3}{pgrey}  \mkopenci{(5.98,-2.2)}{pgrey}
\ebar{6.36}{3.5}{1.3}{ccallb}  \mkfillci{(6.36,3.5)}{ccallb}
\ebar{6.74}{12.7}{2.3}{pblue}  \mkfillsq{(6.74,12.7)}{pblue}
\mkopensq{(0.62,-13.6)}{pgrey} \node[anchor=west, font=\scriptsize] at (0.70,-13.6) {USC};
\mkopenci{(1.58,-13.6)}{pgrey} \node[anchor=west, font=\scriptsize] at (1.66,-13.6) {self-check};
\mkfillci{(3.30,-13.6)}{ccallb}\node[anchor=west, font=\scriptsize] at (3.38,-13.6) {\method{}};
\mkfillsq{(4.32,-13.6)}{pblue} \node[anchor=west, font=\scriptsize] at (4.40,-13.6) {judge-select};
\end{tikzpicture}
\caption{Accuracy change over CoT maj@4 with 95\% paired intervals. The gray band runs to the pass@4 ceiling, the most any selector can gain. Self-check
(open circles) falls below zero everywhere and USC (open squares) straddles it.
Both gpt-oss-120b arms clear zero, but their gap widens with difficulty:
judge-select leads \method{} by 0.6, 3.7 and 9.2 points.}
\label{fig:gain}
\end{figure}
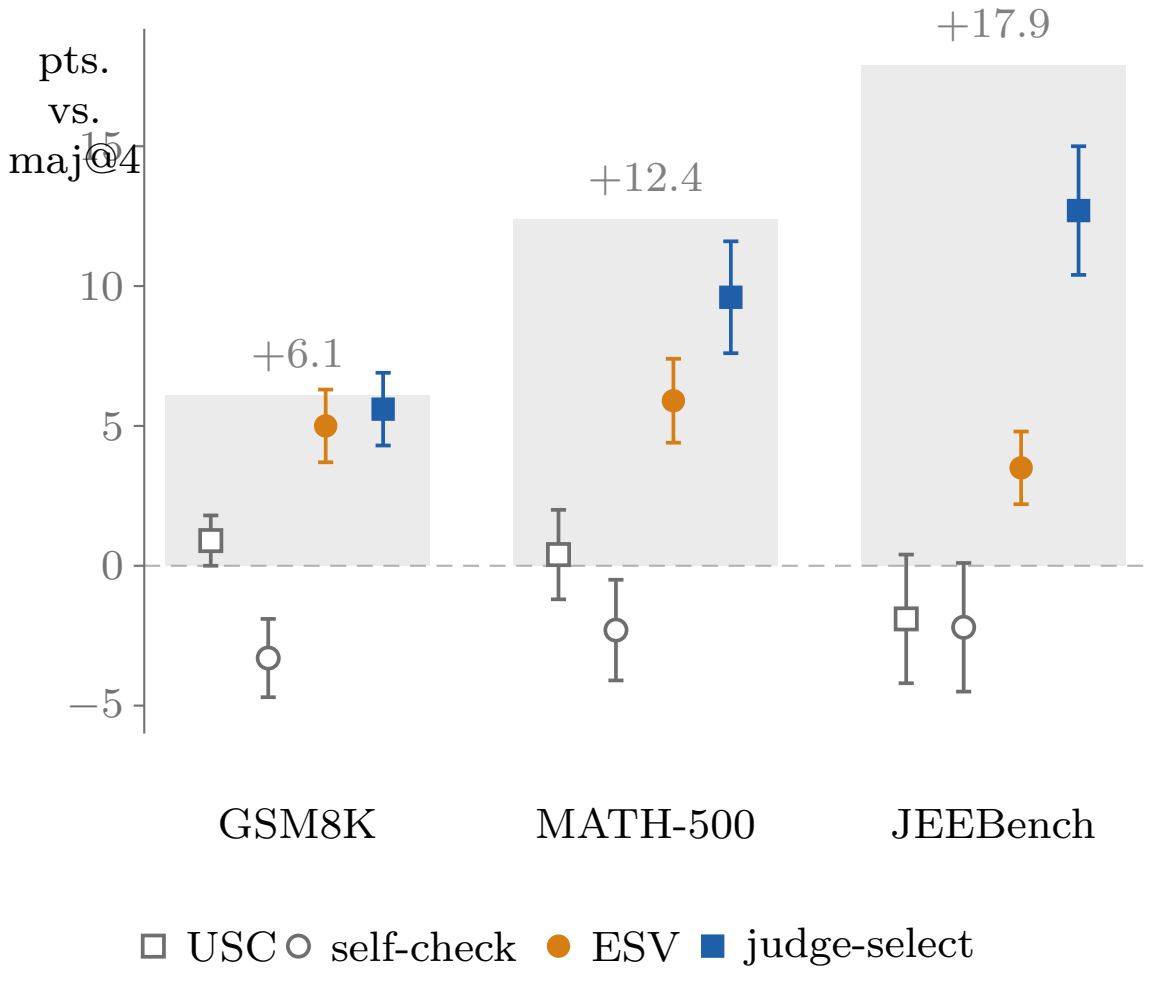

Figure~\ref{fig:trap} shows one accepted strategy-breaking pair and the trap
it manufactures, Figure~\ref{fig:pipeline} draws \method{}'s stages in the
order they run, Figure~\ref{fig:gain} plots the accuracy change of each
selector against CoT maj@4 with paired intervals, and
Table~\ref{tab:methods} states what each selector is permitted to see and do.
All four support the body and are placed here for space.

\begin{figure}[!htbp]
\centering
\begin{tikzpicture}[
  font=\scriptsize,
  base/.style={draw, rounded corners=2.5pt, align=left, inner sep=4pt,
               text width=34mm, anchor=north},
  wide/.style={base, text width=76mm, align=center},
  prob/.style={base, fill=cshared},
  strat/.style={wide, fill=ccode},
  good/.style={base, fill=cout},
  bad/.style={base, fill=ctrap},
  note/.style={wide, fill=white},
  ar/.style={-{Latex[length=2mm]}, semithick},
]
\node[prob] (pl) at (-2.1,0)
  {\textbf{Original $P$}\\[1pt]
   ``In 7 days, Sofie will walk twice as far as \textbf{Brian}.''\\[2pt]
   answer: $35$};
\node[prob] (pr) at (2.1,0)
  {\textbf{Perturbed $P'$}\\[1pt]
   ``In 7 days, Sofie will walk twice as far as \textbf{Brian walks in one
   day}.''\\[2pt]
   answer: $245$};

\node[strat] (s) at (0,-2.25)
  {\textbf{Strategy $S$}, the one $P$ invites:\\[1pt]
   Sofie's 7-day total is $10 \times 7 = 70$; then $70 \div 2$};

\node[good] (tl) at (-2.1,-3.65)
  {$70 \div 2 = 35$\\[1pt]
   $=$ Brian's 7-day total~\yes};
\node[bad] (tr) at (2.1,-3.65)
  {$70 \div 2 = 35$\\[1pt]
   but here $35$ is Brian's \emph{daily} distance~\no};

\node[note] (cc) at (0,-5.45)
  {$S$ omits the final $\times 7$. The trap is $\mathbf{35}$, the original's
   own answer, while $P'$ requires $35 \times 7 = 245$.};

\draw[ar] (pl) -- (s);
\draw[ar] (pr) -- (s);
\draw[ar] (s) -- (tl);
\draw[ar] (s) -- (tr);
\draw[ar, densely dashed] (tr) -- (cc);
\end{tikzpicture}
\caption{A strategy-breaking pair (GSM8K \#131), abridged; both problems also
state that Brian's daily distance is constant. Applying $S$ unchanged to $P'$
runs to completion and returns $35$, so the perturbation turns the original's
correct answer into the trap. The generator and both verifier models
independently returned $245$ for $P'$ and $35$ as the naive answer, and both
verifiers marked $S$ invalid on $P'$. Repeated samples agree on $35$ precisely
because they share the strategy, which is what frequency-based selection cannot
see.}
\label{fig:trap}
\end{figure}
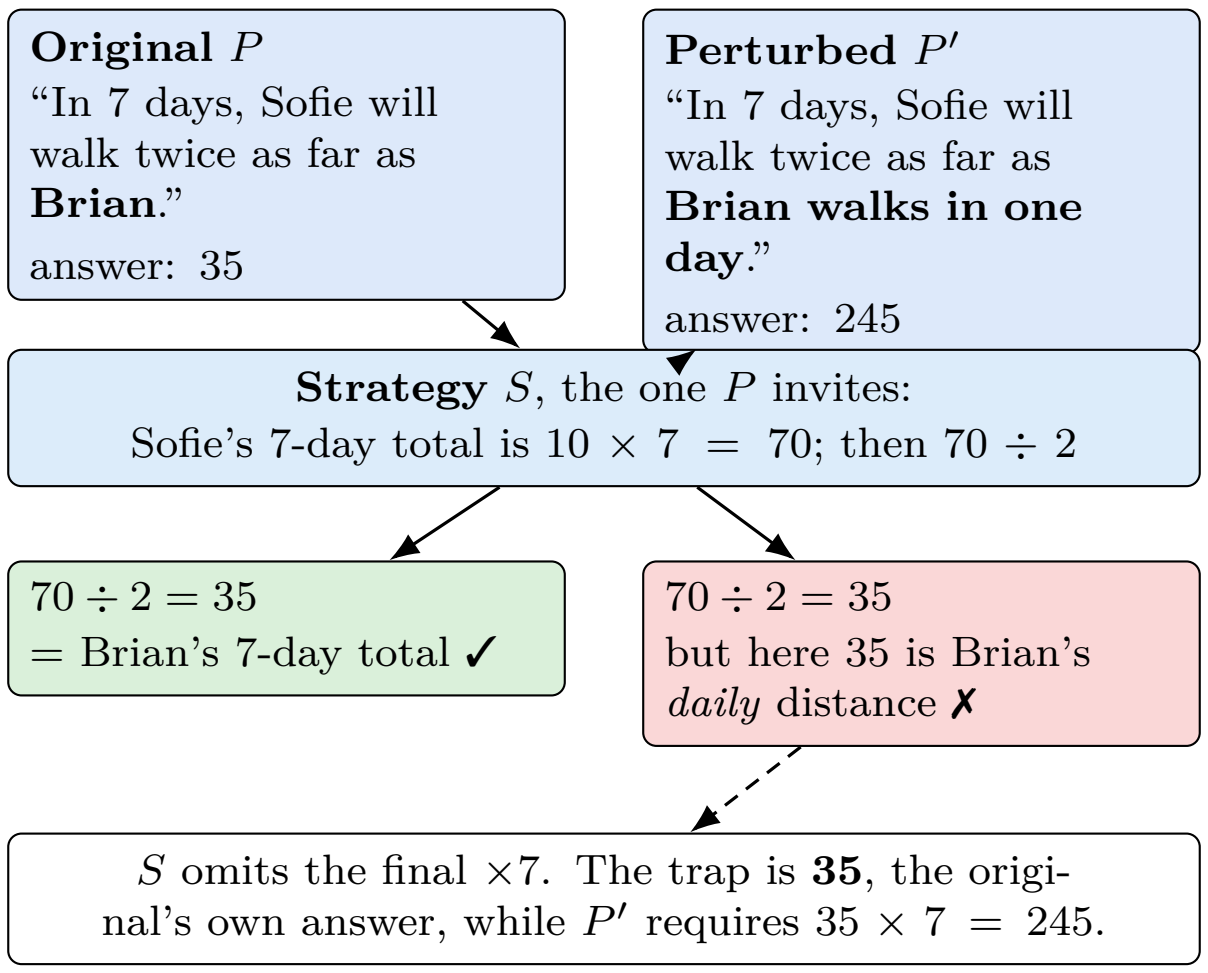

\begin{table}[!htbp]
\centering
\small
\setlength{\tabcolsep}{4pt}
\begin{tabular}{@{}lccccc@{}}
\toprule
 & DQA & maj@4 & USC & Judge & \textbf{\method{}} \\
\midrule
Gold-free             & \yes & \yes & \yes & \yes & \yes \\
Reads shared samples  & \no  & \yes & \yes & \yes & \yes \\
Votes hidden          & \na  & \na  & \no  & \yes & \yes \\
Can overturn $m(x)$   & \na  & \no  & \yes & \yes & \yes \\
May abstain           & \na  & \no  & \no  & \yes & \yes \\
Structured evidence   & \no  & \no  & \no  & \no  & \yes \\
\midrule
Extra calls           & $4^{\dagger}$ & $0$ & $1$ & $1$ & $1$ \\
\bottomrule
\end{tabular}
\caption{What each method may see and do. DQA is Direct QA, Judge the
judge-select control; \emph{structured evidence} means quoted bindings and
checked conditions rather than prose. A dash marks a property that presupposes
selecting among the shared candidates, which Direct QA and CoT maj@4 do not do.
The last two columns differ in one row only, so comparing them isolates
\method{}'s form from the checker's capability. $^{\dagger}$Direct QA draws its
own four samples from a no-reasoning prompt.}
\label{tab:methods}
\end{table}

\subsection{Compute}
\label{app:compute}

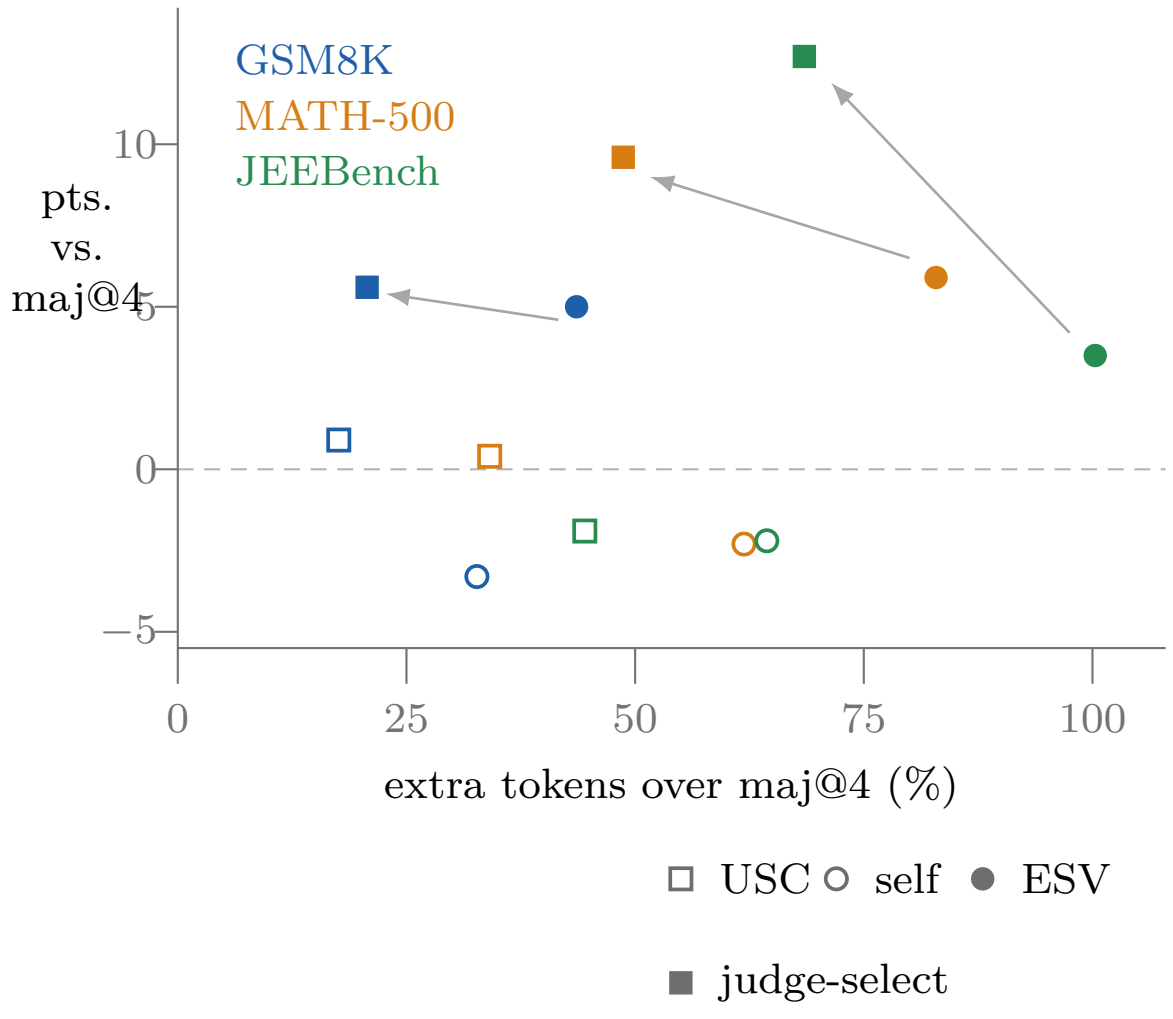
\begin{figure}[t]
\centering
\begin{tikzpicture}[x=0.0605cm, y=0.215cm, font=\footnotesize]
\draw[black!30, densely dashed] (0,0) -- (108,0);
\draw[black!55] (0,-5.5) -- (0,14.2);
\draw[black!55] (0,-5.5) -- (108,-5.5);
\foreach \y in {-5,0,5,10}
  \draw[black!55] (-2.5,\y) -- (0,\y) node[left=0pt, font=\scriptsize] {$\y$};
\foreach \x in {0,25,50,75,100}
  \draw[black!55] (\x,-5.5) -- (\x,-6.6) node[below=0pt, font=\scriptsize] {$\x$};
\node[font=\scriptsize] at (54,-9.8) {extra tokens over maj@4 (\%)};
\node[anchor=south, font=\scriptsize, align=center] at (-11,4.0) {pts.\\vs.\\maj@4};
\draw[-{Latex[length=1.6mm]},black!35,line width=.5pt] (41.6,4.6) -- (22.7,5.4);
\draw[-{Latex[length=1.6mm]},black!35,line width=.5pt] (80.0,6.5) -- (51.6,9.0);
\draw[-{Latex[length=1.6mm]},black!35,line width=.5pt] (97.5,4.2) -- (71.4,11.9);
\mkopensq{(17.6,0.9)}{pblue}  \mkopenci{(32.7,-3.3)}{pblue}
\mkfillci{(43.6,5.0)}{pblue}  \mkfillsq{(20.7,5.6)}{pblue}
\mkopensq{(34.1,0.4)}{ccallb} \mkopenci{(61.9,-2.3)}{ccallb}
\mkfillci{(82.9,5.9)}{ccallb} \mkfillsq{(48.7,9.6)}{ccallb}
\mkopensq{(44.5,-1.9)}{pgreen}\mkopenci{(64.4,-2.2)}{pgreen}
\mkfillci{(100.3,3.5)}{pgreen}\mkfillsq{(68.5,12.7)}{pgreen}
\node[anchor=west, font=\scriptsize, pblue]  at (4,12.6) {GSM8K};
\node[anchor=west, font=\scriptsize, ccallb] at (4,10.9) {MATH-500};
\node[anchor=west, font=\scriptsize, pgreen] at (4,9.2)  {JEEBench};
\mkopensq{(55,-12.6)}{pgrey} \node[anchor=west, font=\scriptsize] at (57,-12.6) {USC};
\mkopenci{(72,-12.6)}{pgrey} \node[anchor=west, font=\scriptsize] at (74,-12.6) {self};
\mkfillci{(88,-12.6)}{pgrey} \node[anchor=west, font=\scriptsize] at (90,-12.6) {\method{}};
\mkfillsq{(55,-15.8)}{pgrey} \node[anchor=west, font=\scriptsize] at (57,-15.8) {judge-select};
\end{tikzpicture}
\caption{Accuracy gain against extra tokens, both over CoT maj@4, with color
for the benchmark and shape for the method. Judge-select sits up and to the left
of \method{} on all three (arrows): cheaper and more accurate. Every method
issues the same single extra call, so the spread along the $x$ axis is reply length alone: \method{}'s per-candidate form costs 1.5 to 2.1 times the
checker tokens free-form adjudication does.}
\label{fig:cost}
\end{figure}

Table~\ref{tab:compute} gives the cost of every method per instance and
Figure~\ref{fig:cost} plots what that cost buys. Both are read off the
\texttt{usage} and latency fields stored on each call, so they reproduce offline
from the records.

\begin{table*}[!htbp]
\centering
\small
\setlength{\tabcolsep}{3pt}
\begin{tabular}{llrrrr}
\toprule
 & Method & Tokens & vs.\ maj@4 & Crit.\ path (s) & Tok./net fix \\
\midrule
GSM8K & CoT maj@4 & 2{,}379 & -- & 6.5 & -- \\
 & USC & 2{,}799 & +17.6\% & 6.7 & 49.1k \\
 & ESV, self & 3{,}158 & +32.7\% & 10.8 & no gain \\
 & Judge-select & 2{,}871 & +20.7\% & 10.0 & \textbf{8.7k} \\
 & ESV, 120b & 3{,}417 & +43.6\% & 13.7 & 20.6k \\
\midrule
MATH-500 & CoT maj@4 & 4{,}592 & -- & 31.8 & -- \\
 & USC & 6{,}159 & +34.1\% & 32.6 & 382k \\
 & ESV, self & 7{,}433 & +61.9\% & 55.6 & no gain \\
 & Judge-select & 6{,}826 & +48.7\% & 44.7 & \textbf{23.2k} \\
 & ESV, 120b & 8{,}399 & +82.9\% & 72.9 & 64.1k \\
\midrule
JEEBench & CoT maj@4 & 7{,}548 & -- & 31.4 & -- \\
 & USC & 10{,}910 & +44.5\% & 32.2 & no gain \\
 & ESV, self & 12{,}411 & +64.4\% & 55.6 & no gain \\
 & Judge-select & 12{,}718 & +68.5\% & 73.0 & \textbf{40.7k} \\
 & ESV, 120b & 15{,}116 & +100.3\% & 92.0 & 216k \\
\bottomrule
\end{tabular}
\caption{Cost per instance. Tokens are prompt plus completion summed over solver and checker; the critical path is the latency a user waits, since the four samples run concurrently. Tokens per net fix is extra tokens times $N$ divided by $F-B$. Solver and checker tokens are counted at equal weight.}
\label{tab:compute}
\end{table*}

\paragraph{What more samples would buy.} We could not run maj@8, but the four
stored samples bound the marginal value of a fifth through eighth. Scoring every
subset of size $k$ and averaging gives, for $k = 1 \dots 4$, maj@$k$ of 85.1,
84.7, 87.9 and 88.5\% on GSM8K, 51.6, 51.8, 54.9 and 56.4\% on MATH-500, and
20.5, 21.0, 23.1 and 24.0\% on JEEBench, reproducing each benchmark's reported
maj@4 exactly. The step from three samples to four is already worth only $+0.60$,
$+1.41$ and $+0.91$ points, and every curve is concave, so four further samples
would have to reverse a decelerating trend to match \method{}'s $+5.0$, $+5.9$
and $+3.5$ or judge-select's $+5.6$, $+9.6$ and $+12.7$. This is an
extrapolation from the samples we have, not a measurement of maj@8, and we
report it as such.

The gate fires on 25.9\%, 61.6\% and 89.9\% of instances. A verification call on a GSM8K instance that reaches it costs 4{,}008 tokens and 27.8 seconds; amortized over all instances it costs 1{,}038 tokens and 7.2 seconds. The gate also directs spending where the errors are: it fires on 45.1\% of breaking $P'$ against 17.5\% of preserving $P'$. Counting checker tokens at $k$ times the price of solver tokens, ESV costs less per net fix than USC for $k < 2.4$ on GSM8K and $k < 6.0$ on MATH-500. USC's total gain is capped at 10 net fixes on GSM8K against ESV's 59, so additional USC spending cannot reach ESV's accuracy.

\subsection{Large solvers}
\label{app:large}

With Qwen2.5-72B-Instruct, Mistral-Small-3.2-24B and Llama-3.3-70B as solvers on GSM8K (1{,}170 instances), CoT maj@4 reaches 96.9\% (96.9, 96.7 and 97.2\% per solver) and pass@4 99.0\%. Only 88 instances disagree, 24 of them fixable, and one of the 1{,}082 unanimous instances is wrong. With 2.1 points of headroom there is little for any selector to recover, so we report this arm as the reason the small solvers are the test and did not run verification on it.

\subsection{Verification transcripts}
\label{app:transcripts}

Both examples come from the self-check arm on the GSM8K development split.

\noindent\fbox{\parbox{0.95\columnwidth}{\small
\textbf{Pair 4 (breaking).} The perturbed problem pays Sara nothing for her first four mowings. The checker quotes the clause ``\$4 for each time she mows the lawn after the first 4 times'' word for word, then marks the trap solution's binding $4x$, which pays for every mowing, as \textsc{matches} and returns \textsc{sound}.}}

\vspace{4pt}
\noindent\fbox{\parbox{0.95\columnwidth}{\small
\textbf{Pair 36 (breaking).} The first leg of a relay is shortened to 380\,m while the nominal times stay unchanged. The checker writes that the first runner runs 380\,m in 60\,s and marks the condition \textsc{holds}, calls the trap solution's per-meter rescaling correct, and argues for the trap in its divergence analysis.}}

\subsection{Prompts and decoding settings}
\label{app:prompts}

The claim under test in Section~\ref{sec:ablation} is a claim about a prompt, so
we give the two prompts it compares in full. Every other prompt (CoT, Direct QA,
USC, the generator and the two pair verifiers) is in
\texttt{method\_prompts.py} and \texttt{perturb\_prompts.py} in the release of
Appendix~\ref{app:code}, which is the same text the stored responses were drawn
with. Both prompts below are shown with the per-candidate tag list abbreviated
to one candidate; the code emits the same block for each.

\paragraph{Decoding.} Solvers draw four samples at temperature $0.7$ with seeds
$0$--$3$ and a 4{,}000-token cap (6{,}000 on MATH-500). The checker,
judge-select and USC run at temperature $0$ with gpt-oss-120b at reasoning effort
medium. The pair generator (DeepSeek-V4-Flash) runs at reasoning effort high
with a 16{,}000-token cap, and both pair verifiers (Qwen3.5-35B-A3B,
gpt-oss-120b) at effort medium with a 12{,}000-token cap; all three send no
temperature and take the provider default. No other sampling parameter is
sent on any call, so \texttt{top\_p}, \texttt{top\_k} and the repetition
penalties also take the provider default; what a reproduction must send is
therefore the absence of these fields, not a value for them. Every request
pins OpenRouter's provider routing to \texttt{sort: price}, and for every model
not served natively in 4 bits it also sends an explicit quantization allowlist
(int8, fp6, fp8, fp16, bf16, fp32, unknown) that excludes 4-bit serving. Each
stored response records the identifier requested, the token cap, the
temperature and seed sent, the finish reason and the token counts, so the
decoding applied to any individual call can be read off the release; the
serving provider OpenRouter resolved to is constrained by the routing
preferences but is not itself recorded. The three datasets were frozen on
2026-11-01 (JEEBench), 2026-11-04 (GSM8K) and 2026-11-09 (MATH-500), and the
served identifiers are the dated endpoint records from that window. A reply
truncated at the cap is retried once with the cap doubled, and a pair is
rejected if a verifier's confidence falls below $0.6$. Solutions shown to the
checker are clipped to 6{,}000 characters.

\paragraph{ESV stage-2 verification prompt.}
\begin{lstlisting}[basicstyle=\scriptsize\ttfamily,numbers=none,xleftmargin=0pt,frame=single,breaklines=true,columns=fullflexible]
You are performing Explicit Strategy Verification (ESV)
on several solutions to the same problem: stage 2
(strategy verification), stage 3 (failure recognition)
and stage 4 (diagnosis).
The solutions reach DIFFERENT final answers, so at most
one of them is correct, and all of them may be wrong.
They are listed in arbitrary order; how many attempts
reached each answer is not shown and is not evidence.

For EACH solution:
1. strategy: the method it uses, in one sentence.
2. bindings: read the PROBLEM STATEMENT itself (not the
solution's restatement of it) and quote, word for word,
every given the method depends on: definitions and labels
of variables, points and sides, ranges and intervals,
constraints, special conditions, and exactly what is
asked. For each quote, say how this solution used it and
mark MATCHES (used exactly as stated) or MISMATCH
(misread, replaced by a familiar convention, dropped or
changed).
3. conditions: each condition the method needs
(hypotheses of each theorem or formula, which case
applies, domains, signs), marked HOLDS or VIOLATED for
THIS problem.
Then divergence: find the first step where the solutions
disagree, quote each version, and say which one the
problem statement supports and why.
Then, for EACH solution, the verdict:
- SOUND only if every binding MATCHES, every condition
HOLDS, and recomputing its key steps finds no error.
- FLAWED otherwise, with failure_type BINDING (a given of
the problem is misused), METHOD (the method cannot work
for this problem) or COMPUTATION (only a calculation or
algebra error); the first flawed step quoted word for
word from that solution; and the corrected fact (what
that step should say).
For a SOUND solution write NONE in failure_type,
flawed_step and corrected_fact.

{problem}  {format note}

Solution i:
<reasoning>...</reasoning>
<final_answer>...</final_answer>

Format rules (strict):
- Output ONLY the tags listed, in order, each with an
opening and a closing tag. No JSON, no Markdown fences,
no text outside the tags.
- Write LaTeX normally with single backslashes; do not
escape anything.
- Be concise: about 150 words per solution in total, one
short line per check, and never leave a tag unfinished.

Output exactly these tags:
<s1_strategy>Solution 1: the method this solution uses,
in one sentence</s1_strategy>
<s1_bindings>Solution 1: one line per given of the
problem the method depends on: "quote from the problem"
-> how this solution used it -> MATCHES or MISMATCH
</s1_bindings>
<s1_conditions>Solution 1: one line per condition the
method needs: condition - HOLDS or VIOLATED in THIS
problem - one-line reason</s1_conditions>
<divergence>the first step where the solutions disagree:
quote each version and say which one the problem
statement supports, and why</divergence>
<s1_verdict>Solution 1: SOUND or FLAWED</s1_verdict>
<s1_failure_type>Solution 1: BINDING, METHOD or
COMPUTATION; NONE if SOUND</s1_failure_type>
<s1_flawed_step>Solution 1: the first flawed step, quoted
word for word from the solution; NONE if SOUND
</s1_flawed_step>
<s1_corrected_fact>Solution 1: what that step should say
instead; NONE if SOUND</s1_corrected_fact>
\end{lstlisting}

\paragraph{Judge-select control prompt.}
\begin{lstlisting}[basicstyle=\scriptsize\ttfamily,numbers=none,xleftmargin=0pt,frame=single,breaklines=true,columns=fullflexible]
You are judging several solutions to the same problem.
The solutions reach DIFFERENT final answers, so at most
one of them is correct, and all of them may be wrong.
They are listed in arbitrary order; how many attempts
reached each answer is not shown and is not evidence.

Work out which solution, if any, is correct, and select
it. If no solution is correct, select NONE.

{problem}  {format note}

Solution i:
<reasoning>...</reasoning>
<final_answer>...</final_answer>

Format rules (strict):
- Output ONLY the tags listed, in order, each with an
opening and a closing tag. No JSON, no Markdown fences,
no text outside the tags.
- Write LaTeX normally with single backslashes; do not
escape anything.
- Be concise: about 150 words per solution in the
reasoning tag, and never leave a tag unfinished.
- Identify your selection by its Solution LABEL, the
integer 1-n printed above it, or the word NONE. Do not
write the answer itself: <selected_solution>2</selected_
solution> means Solution 2, not the answer 2.

Output exactly these tags:
<reasoning>how you decided, in your own words</reasoning>
<selected_solution>ONE integer from 1 to n: the label of
the solution you select, NOT its answer value; or NONE if
no solution is correct</selected_solution>
\end{lstlisting}

\vspace{4pt}
\noindent The two differ in form and in nothing else: same checker, same
candidates in the same order, same hidden vote counts, same licence to reject
every candidate, and a free-form reasoning tag first in both so the comparison
is structured against unstructured reasoning rather than reasoning against none.

\subsection{Prompt note}

During development the CoT and Direct QA prompts told every model to answer multiple-choice questions with an option letter, and Gemma-3-4B returned a bare letter on 138 of 168 GSM8K samples. The instruction now depends on the answer format (integer, numeric or expression), with the multiple-choice wording unchanged for JEEBench. Every run reported in this paper carries the same prompt build, and no bare option letter survives in any of the 17{,}312 stored GSM8K and MATH-500 samples.